\documentclass{article}

\usepackage[final]{neurips_2022}

\usepackage[utf8]{inputenc} 
\usepackage[T1]{fontenc}    
\usepackage[hidelinks]{hyperref}       
\usepackage{url}            
\usepackage{booktabs}       
\usepackage{amsfonts}       
\usepackage{nicefrac}       
\usepackage{microtype}      
\usepackage{xcolor}         

\title{Monte Carlo Augmented Actor-Critic for Sparse Reward Deep Reinforcement Learning from Suboptimal Demonstrations}

\author{%
  Albert Wilcox \\
  UC Berkeley \\
  \texttt{albertwilcox@berkeley.edu} \\
  \And
  Ashwin Balakrishna \\
  UC Berkeley \\
  \texttt{ashwin\_balakrishna@berkeley.edu} \\
  \And
  Jules Dedieu \\
  UC Berkeley \\
  \texttt{jules\_dedieu@berkeley.edu} \\
  \And
  Wyame Benslimane \\
  UC Berkeley \\
  \texttt{wyame.benslimane@berkeley.edu} \\
  \And
  Daniel S. Brown \\
  University of Utah \\
  \texttt{dsbrown@cs.utah.edu} \\
  \And
  Ken Goldberg \\
  UC Berkeley \\
  \texttt{goldberg@berkeley.edu} \\
}

\usepackage{microtype}
\usepackage{subfigure}
\usepackage{booktabs} 

\usepackage{hyperref}

\usepackage{graphics}
\usepackage[pdftex]{graphicx}
\usepackage{wrapfig}
\DeclareGraphicsExtensions{.pdf,.png,.jpg}
\usepackage{epsfig}
\usepackage[font={small}]{caption}
\usepackage[rightcaption]{sidecap}
\usepackage{pbox}

\usepackage{bigstrut}
\usepackage{makecell}

\usepackage{mathtools}
\usepackage{amssymb,amsmath}
\usepackage{gensymb} 
\usepackage{nicefrac}       
\numberwithin{equation}{section} 
\usepackage{algorithm}
\usepackage{algpseudocode}

\usepackage{amsthm}

\theoremstyle{definition}
\theoremstyle{assumption}

\DeclareMathOperator*{\argmax}{arg\,max}

\DeclarePairedDelimiter{\abs}{|}{|}
\DeclarePairedDelimiter{\set}{\{}{\}}

\newcommand{\algname}{Monte Carlo augmented Actor-Critic}
\newcommand{\algabbr}{MCAC}
\newcommand{\new}[1]{\textcolor{black}{#1}}
\begin{document}

\maketitle

\begin{abstract}
Providing densely shaped reward functions for RL algorithms is often exceedingly challenging, motivating the development of RL algorithms that can learn from easier-to-specify sparse reward functions. This sparsity poses new exploration challenges. One common way to address this problem is using demonstrations to provide initial signal about regions of the state space with high rewards. However, prior RL from demonstrations algorithms introduce significant complexity and many hyperparameters, making them hard to implement and tune. We introduce \algname{} (\algabbr{}), a parameter free modification to standard actor-critic algorithms which initializes the replay buffer with demonstrations and computes a modified $Q$-value by taking the maximum of the standard temporal distance (TD) target and a Monte Carlo estimate of the reward-to-go. This encourages exploration in the neighborhood of high-performing trajectories by encouraging high $Q$-values in corresponding regions of the state space. Experiments across $5$ continuous control domains suggest that \algabbr{} can be used to significantly increase learning efficiency across $6$ commonly used RL and RL-from-demonstrations algorithms. See \url{https://sites.google.com/view/mcac-rl} for code and supplementary material.
\end{abstract}

\section{Introduction}
Reinforcement learning has been successful in learning complex skills in many environments \citep{mnih2015human, silver2016mastering, schulman2017proximal}, but providing dense, informative reward functions for RL agents is often very challenging~\citep{xie2019deep,krakovna2020specification,wu2020learning}. This is particularly challenging for high-dimensional control tasks, in which there may be a large number of factors that influence the agent's objective. In many settings, it may be much easier to provide sparse reward signals that simply convey high-level information about task progress, such as whether an agent has completed a task or has violated a constraint. However, optimizing RL policies given such reward signals can be exceedingly challenging, as sparse reward functions may not be able to meaningfully distinguish between a wide range of different policies.

This issue can be mitigated by leveraging demonstrations, which provide initial signal about desired behaviors. Though demonstrations may often be suboptimal in practice, they should still serve to encourage exploration in promising regions of the state space, while allowing the agent to explore behaviors which may outperform the demonstrations. Prior work has considered a number of ways to leverage demonstrations to improve learning efficiency for reinforcement learning, including by initializing the policy to match the behavior of the demonstrator~\citep{rajeswaran2018learning,peng2019advantageweighted}, using demonstrations to explicitly constrain agent exploration~\citep{thananjeyan2020safety,thananjeyan2021recovery, LS3}, and  introducing auxiliary losses to incorporate demonstration data into policy updates~\citep{nair2018overcoming,hester2018deep,gao2018reinforcement}. While these algorithms have shown impressive performance in improving the sample efficiency of RL algorithms, they add significant complexity and hyperparameters, making them difficult to implement and tune for different tasks.

We present \algname{} (\algabbr{)}, which introduces an easy-to-implement, but highly effective, change that can be readily applied to existing actor-critic algorithms without the introduction of any additional hyperparameters and only minimal additional complexity. The idea is to encourage initial optimism in the neighborhood of successful trajectories, and progressively reduce this optimism during learning so that it can continue to explore new behaviors. To operationalize this idea, \algabbr{} introduces two modifications to existing actor-critic algorithms. First, \algabbr{} initializes the replay buffer with task demonstrations. Second, \algabbr{} computes a modified target $Q$-value for critic updates by taking the maximum of the standard temporal distance targets used in existing actor critic algorithms and a Monte Carlo estimate of the reward-to-go. The intuition is that Monte Carlo value estimates can more effectively capture longer-term reward information, making it possible to rapidly propagate reward information from demonstrations through the learned $Q$-function. This makes it possible to prevent underestimation of values in high-performing trajectories early in learning, as high rewards obtained later in a trajectory may be difficult to initially propagate back to earlier states with purely temporal distance targets~\citep{wright2013exploiting}.
Experiments on five continuous control domains suggest that \algabbr{} is able to substantially accelerate exploration for both standard RL algorithms and recent RL from demonstrations algorithms in sparse reward tasks.

\section{Related Work}
\subsection{Reinforcement Learning from Demonstrations}

One standard approach for using demonstrations for RL first uses imitation learning~\citep{argall2009survey} to pre-train a policy, and then fine-tunes this policy with on-policy reinforcement learning algorithms \citep{schaal1997learning,kober2014policy,peng2019advantageweighted, rajeswaran2018learning}. However, initializing with suboptimal demonstrations can hinder learning and using demonstrations to initialize only a policy is inefficient, since they can also be used for $Q$-value estimation. 

Other approaches leverage demonstrations to explicitly constrain agent exploration. \citet{thananjeyan2020safety} and \citet{LS3} propose model-based RL approaches that uses suboptimal demonstrations to iteratively improve performance by ensuring consistent task completion during learning. Similarly, \citet{jing2020reinforcement} also uses suboptimal demonstrations to formulate a soft-constraint on exploration. However, a challenge with these approaches is that they introduce substantial algorithm complexity, making it difficult to tune and utilize these algorithms in practice. For example, while~\citet{thananjeyan2020safety} does enable iterative improvement upon suboptimal demonstrations, they require learning a model of system dynamics and a density estimator to capture the support of successful trajectories making it challenging to scale to high-dimensional observations. 

Finally, many methods introduce auxiliary losses to incorporate demonstrations into policy updates~\citep{kim2013learning,gao2018reinforcement,kang2018policy}. Deep Deterministic Policy Gradients from Demonstrations (DDPGfD)~\citep{vecerik2018leveraging} maintains all the demonstrations in a separate replay buffer and uses prioritized replay to allow reward information to propagate more efficiently. \citet{nair2018overcoming}, and \citet{hester2018deep} use similar approaches, where demonstrations are maintained separately from the standard replay buffer and additional policy losses encourage imitating the behavior in the demonstrations. Meanwhile, RL algorithms such as AWAC~\citep{AWAC} pretrain using demonstration data to constrain the distribution of actions selected during online exploration. While these methods often work well in practice, they often increase algorithmic complexity and introduce several additional hyperparameters that are difficult and time consuming to tune. By contrast, \algabbr{} does not increase algorithmic complexity, is parameter-free, and can easily be wrapped around any existing actor-critic algorithm.

\subsection{Improving $Q$-Value Estimates}
The core contribution of this work is an easy-to-implement, yet highly effective, method for stabilizing actor-critic methods for sparse reward tasks using demonstrations and an augmented $Q$-value target. There has been substantial literature investigating learning stability challenges in off-policy deep $Q$-learning and actor-critic algorithms. See~\citet{deadlytriad} for a more thorough treatment of the learning stability challenges introduced by combining function approximation, bootstrapping, and off-policy learning, as well as prior work focused on mitigating these issues. 

One class of approaches focuses on developing new ways to compute target $Q$-values~\citep{konidaris2011td_gamma,wright2013exploiting,double-dqn,GAE}.~\citet{double-dqn} computes target $Q$-values with two $Q$-networks, using one to select actions and the other to measure the value of selected actions, which helps to prevent the $Q$-value over-estimation commonly observed in practice in many practical applications of $Q$-learning.
TD3~\citep{fujimoto2018addressing} attempts to address overestimation errors by taking the minimum of two separate $Q$-value estimates, but this can result in underestimation of the true $Q$-value target. \citet{kuznetsov2020controlling} uses an ensemble of critics to adaptively address estimation errors in $Q$-value targets, but introduces a number of hyperparameters which must be tuned separately for different tasks. \citet{bellemare2016unifying} and \cite{ostrovski2017count} consider a linear combination of a TD-1 target and Monte Carlo target. \citet{wright2015cfqi, GAE} and \citet{multi-step} consider a number of different estimators of a policy's value via $n$-step returns, which compute $Q$-targets using trajectories with $n$ contiguous transitions followed by a terminal evaluation of the $Q$-value after $n$ steps. Each of these targets make different bias and variance tradeoffs that can affect learning dynamics.
 
Similar to our work, \citet{wright2013exploiting} explore the idea of taking a maximum over a bootstrapped target $Q$-value (TD-1 target) and a Monte Carlo estimate of the return-to-go to improve fitted $Q$-iteration. However, \citet{wright2013exploiting} focuses on fully offline $Q$-learning and only considers simple low-dimensional control tasks using $Q$-learning with linear value approximation. \new{There are numerous reasons to extend these ideas to online deep RL. First, deep RL algorithms are often unstable, and using the ideas in \citet{wright2013exploiting} to improve $Q$ estimates is a promising way to alleviate this, as we empirically verify. Second, while offline learning has many important applications, online RL is far more widely studied, and we believe it is useful to study the effects these ideas have in this setting.}  To the best of our knowledge, \algabbr{} is the first application of these ideas to online actor-critic algorithms with deep function approximation\new{, and w}e find that \new{it} yields surprising improvements in RL performance on complex high-dimensional continuous control tasks. 

\section{Problem Statement}
\label{sec:problem}
We consider a Markov Decision Process (MDP) described by a tuple $(\mathcal{S}, \mathcal{A}, p, r, \gamma, T)$ with a state set $\mathcal{S}$, an action set $\mathcal{A}$, a transition probability function $p: \mathcal{S} \times \mathcal{A} \times \mathcal{S} \rightarrow [0, 1]$, a reward function $r: \mathcal{S} \times \mathcal{A} \rightarrow \mathbb{R}$, a discount factor $\gamma$, and finite time horizon $T$. In each state $s_t \in \mathcal{S}$ the agent chooses an action $a_t \in \mathcal{A}$ and observes the next state $s_{t+1}\sim p(\cdot | s_t, a_t)$ and a reward $r(s_t, a_t) \in \mathbb{R}$. The agent acts according to a policy $\pi$, which induces a probability distribution over $\mathcal{A}$ given the state, $\pi(a_t|s_t)$. The agent's goal is to find the policy $\pi^*$ which at any given $s_t \in \mathcal{S}$ maximizes the expected discounted sum of rewards, 
\begin{equation}
    \pi^* = \argmax_\pi \mathbb{E}_{\tau \sim \pi}\left[ \sum^T_{t=0} \gamma^t r(s_t, a_t) \right],
\end{equation}
where $\tau =(s_0, a_0, s_1, a_1, \ldots s_T)$ and $\tau \sim \pi$ indicates the distribution of trajectories induced by evaluating policy $\pi$ in the MDP.

We make additional assumptions specific to the class of problems we study. First, we  assume that all transitions in the replay buffer are elements of complete trajectories; this is a reasonable assumption as long as all transitions are collected from rolling out some policy in the MDP. Second, we assume the agent has access to an offline dataset $\mathcal{D}_\mathrm{offline}$ of (possibly suboptimal) task demonstrations. Finally, we focus on settings where the reward function is sparse in the sense that most state transitions do not induce a change in reward, causing significant exploration challenges for RL algorithms.


\section{Preliminaries: Actor-Critic Algorithms}
\label{sec:prelims}
For a given policy $\pi$, its state-action value function $Q^\pi$ is defined as 
\begin{equation}\label{eq:q-def}
    Q^\pi(s_t, a_t) = \mathbb{E}_{\tau \sim \pi}\left[\sum^T_{k=t} \gamma^{k-t}r(s_k, a_k) \right].
\end{equation}

Actor-critic algorithms learn a sample-based approximation to $Q^\pi$, denoted $Q^\pi_\theta$, and a policy $\pi_\phi$ which selects actions to maximize $Q_\theta^\pi$, with a function approximator (typically a neural network) parameterized by $\theta$ and $\phi$ respectively. During the learning process, they alternate between regressing $Q^\pi_\theta$ to predict $Q^\pi$ and optimizing $\pi_\phi$ to select actions with high values under $Q^\pi_\theta$.

Exactly computing $Q^\pi$ targets to train $Q^\pi_\theta$ is typically intractable for arbitrary policies in continuous MDPs, motivating other methods for estimating them. One such method is to simply collect trajectories $(s_t, a_t, r_t, s_{t+1}, \hdots s_{T-1}, a_{T-1}, r_{T-1}, s_T)$ by executing the learned policy $\pi_\phi$ from state $s_t$, computing a Monte Carlo estimate of the reward-to-go defined as follows:
\begin{equation}\label{eq:mc-target}
    Q^{\textrm{target}}_\textrm{MC}(s_t, a_t) = \sum^T_{k=t} \gamma^{k-t}r(s_k, a_k),
\end{equation}
and fitting $Q^\pi_\theta$ to these targets. 

However, $Q^{\textrm{target}}_\textrm{MC}$ can be a high variance estimator of the reward-to-go~\citep{GAE,sutton2018reinforcement}, motivating the popular one-step temporal difference target (TD-1 target) to help stabilize learning:  
\begin{equation}\label{eq:td-target}
    Q^{\textrm{target}}_\textrm{TD}(s_t, a_t) = r(s_t, a_t) + \gamma Q^\pi_{\theta'}(s_{t+1}, a_{t+1}),
\end{equation}
where $a_{t+1} \sim \pi_\phi(s_{t+1})$, which is recursively defined based on only a single $(s_t, a_t, s_{t+1}, r_t)$ transition. Here $\theta'$ is the parameters of a lagged target network as in~\citep{double-dqn}. There has also been interest in computing TD-$n$ targets, which instead sum rewards for $n$ timesteps and then use $Q$-values from step $n+1$~\citep{wright2015cfqi, GAE, multi-step}.

\section{\algname{}}
\label{sec:method}

\subsection{\algabbr{} Algorithm}
\label{sec:mcaqi}
The objective of \algabbr{} is to efficiently convey information about sparse rewards from suboptimal demonstrations to $Q^\pi_\theta$ in order to accelerate policy learning while still maintaining learning stability. To do this, \algabbr{} combines two different methods of computing targets for fitting $Q$-functions to enable efficient value propagation throughout the state-action space while also learning a $Q$-value estimator with low enough variance for stable learning. To operationalize this idea, \algabbr{} defines a new $Q$-function target for training $Q^\pi_\theta$ by taking the maximum of the Monte Carlo $Q$-target (eq~\ref{eq:mc-target}) and the temporal difference $Q$-target (eq~\ref{eq:td-target}): $\max\big[Q^{\textrm{target}}_\textrm{TD}(s_t, a_t), Q^{\textrm{target}}_\mathrm{MC}(s_t, a_t)\big]$.

The idea here is that early in learning, a $Q$-function trained only with temporal difference targets will have very low values throughout the state-action space as it may be very difficult to propagate information about delayed rewards through the temporal difference target for long-horizon tasks with sparse rewards~\citep{deadlytriad}. On the other hand, the Monte Carlo $Q$-target can easily capture long-term rewards, but often dramatically underestimates $Q$-values for poorly performing trajectories~\citep{wright2013exploiting}.
Thus, taking a maximum over these two targets serves to initially boost the $Q$-values for transitions near high performing trajectories while limiting the influence of underestimates from the Monte Carlo estimate.

\algabbr{} can also be viewed as a convenient way to balance the bias and variance properties of the Monte Carlo and temporal difference $Q$-targets. The Monte Carlo $Q$-target is well known to be an unbiased estimator of policy return, but have high variance~\citep{wright2013exploiting}. Conversely, temporal difference targets are known to be biased, but have much lower variance~\citep{wright2013exploiting, double-dqn}. Thus, as the temporal difference target is typically negatively biased (an underestimate of the true policy return) on successful trajectories early in learning due to the challenge of effective value propagation, computing the maximum of the temporal difference and Monte Carlo targets helps push the \algabbr{} target value closer to the true policy return. Conversely, the Monte Carlo targets are typically pessimistic on unsuccessful trajectories, and their high variance makes it difficult for them to generalize sufficiently to distinguish between unsuccessful trajectories that are close to being successful and those that are not. Thus, computing the maximum of the temporal difference and Monte Carlo targets also helps to prevent excessive pessimism in evaluating unsuccessful trajectories. Notably, \algabbr{} does not constrain its $Q$-targets explicitly based on the transitions in the demonstrations, making it possible for the policy to discover higher performing behaviors than those in the demonstrations as demonstrated in Section~\ref{sec:exps}.

\subsection{\algabbr{} Practical Implementation}\label{sec:mcaac}

\begin{algorithm} [t]
\caption{\algname{}}\label{alg:mcac}
\begin{algorithmic}[1]
\Require Offline dataset $\mathcal{D}_\mathrm{offline}$.
\Require Total training episodes $N$, batch size $M$, Pretraining Steps $N_p$
\Require Episode time horizon $T$.
\State Initialize replay buffer $\mathcal{R} := \mathcal{D_\mathrm{offline}}$.
\State Initialize agent $\pi_\phi$ and critic $Q^\pi_\theta$ using data from $\mathcal{D_\mathrm{offline}}$.
\For {$i \in \set{1, \ldots, N_p}$}
    \State Sample $\mathcal{B} \subsetneq \mathcal{R}$ such that $\abs{\mathcal{B}} = M$.
    \State  Optimize $Q_\theta^\pi$ on $\mathcal{B}$ to minimize loss in eq~\eqref{eq:mcac_loss}. Optimize policy $\pi_\phi$ to maximize $Q_\theta$.
\EndFor
\For {$i \in \set{1, \ldots, N}$}
    \State Initialize episode buffer $\mathcal{E} = \set{}$.
    \State Observe state $s_1^i$.
    \For{$j \in \set{1, \ldots, T}$}
        \State Sample and execute $a_t^i \sim \pi_\theta(s_j^i)$, observing $s_{j+1}^i, r_j^i$.
        \State $\tau_j^i \gets (s_j^i, a_j^i, s_{j+1}^i, r_j^i)$
        \State $\mathcal{E} \gets \mathcal{E} \cup \set{\tau_j^i}$.
        \State Sample $\mathcal{B} \subsetneq \mathcal{R}$ such that $\abs{\mathcal{B}} = M$.
        \State  Optimize $Q_\theta^\pi$ on $\mathcal{B}$ to minimize loss in eq~\eqref{eq:mcac_loss}. Optimize policy $\pi_\phi$ to maximize $Q_\theta$.
    \EndFor
    \For{$\tau_j^i \in \mathcal{E}$}
        \State Compute $Q^{\textrm{target}}_\textrm{MC-$\infty$}(s_j^i, a_j^i)$ as in eq~\eqref{eq:mc-target-infinite}.
        \State $\tau_j^i \gets (s_j^i, a_j^i, s_{j+1}^i, r_j^i, Q^{\textrm{target}}_\textrm{MC-$\infty$}(s_j^i, a_j^i))$.
    \EndFor
    \State $\mathcal{R} \gets \mathcal{R} \cup \mathcal{E}$.
\EndFor
\end{algorithmic}
\end{algorithm}

\algabbr{} can be implemented as a wrapper around any actor-critic RL algorithm; we consider 6 options in experiments (Section~\ref{sec:exps}). \algabbr{} starts with an offline dataset of suboptimal demonstrations $\mathcal{D}_\mathrm{offline}$, which are used to initialize a replay buffer $\mathcal{R}$. Then, during each episode $i$, we collect a full trajectory $\tau^i$, where the $j^{\text{th}}$ transition $(s^i_j, a^i_j, s^i_{j+1}, r^i_j)$ in $\tau^i$ is denoted by $\tau^i_j$.

Next, consider any actor-critic method using a learned $Q$ function approximator $Q_\theta(s_t, a_t)$ that, for a given transition $\tau^i_j = (s_j^i, a_j^i, s_{j+1}^i, r_j^i) \in \tau^i \subsetneq \mathcal{R}$ is updated by minimizing the following loss:
\begin{equation}
    \label{eq:vanilla_rl_loss}
    J(\theta) = \ell\left(Q_\theta(s^i_j, a^i_j), Q^\textrm{target}(s^i_j, a^i_j)\right),
\end{equation}
where $\ell$ is an arbitrary differentiable loss function and $Q^\textrm{target}$ is the target value for regressing $Q_\theta$. We note that $Q^\textrm{target}$ is defined by the choice of actor-critic method. To implement \algabbr{}, we first calibrate the Monte Carlo targets with temporal difference targets (which provide infinite-horizon $Q$ estimates) by computing the infinite horizon analogue of the Monte Carlo target defined in Equation~\ref{eq:mc-target}, which assumes the last observed reward value will repeat forever and uses this to add an infinite sum of discounted rewards, and is given by
\begin{equation}
    \label{eq:mc-target-infinite}
    Q^{\textrm{target}}_\textrm{MC-$\infty$}(s_j^i, a_j^i) = \gamma^{T-j+1}\frac{r^i_T}{1-\gamma} + \sum^T_{k=j} \gamma^{k-j}r(s_k^i, a_k^i).
\end{equation}
Then, we simply replace the target with a maximum over the original target and the Monte Carlo target defined in Equation~\ref{eq:mc-target-infinite}, given by
\begin{equation}
    Q^\textrm{target}_\textrm{\algabbr{}}(s_j^i, a_j^i) = \max\big[Q^\textrm{target}(s^i_j, a^i_j), Q^{\textrm{target}}_\textrm{MC-$\infty$}(s^i_j, a^i_j)\big].
\end{equation}
This results in the following loss function for training $Q_\theta$:
\begin{equation}
    \label{eq:mcac_loss}
    J(\theta) = \ell\left(Q_\theta(s^i_j, a^i_j),  Q^\textrm{target}_\textrm{\algabbr{}}(s_j^i, a_j^i)\right).
\end{equation}


The full \algabbr{} training procedure (Algorithm~\ref{alg:mcac}) alternates between updating $Q^\pi_\theta$ using the method described above, followed by optimizing the policy $\pi_\phi$ using any standard policy update method.

\section{Experiments}
\label{sec:exps}
In the following experiments we study (1) whether \algabbr{} enables more efficient learning when built on top of standard actor-critic RL algorithms and (2) whether \algabbr{} can be applied to improve prior algorithms for RL from demonstrations. See the supplementary material for code and instructions on how to run experiments for reproducing all results in the paper and additional experiments studying the impact of demonstration quantity, quality, and other algorithmic choices such as whether to pretrain learned networks on demonstration data before online interaction.

\subsection{Experimental Procedure}\label{sec:metrics}
All experiments were run on a set of 24 Tesla V100 GPUs through a combination of Google Cloud resources and a dedicated lab server. We aggregate statistics over 10 random seeds for all experiments, reporting the mean and standard error across the seeds with exponential smoothing. Details on hyperparameters and implementation details are provided in the supplementary material.

\subsection{Domains}\label{subsec:domains}
\begin{figure}
    \centering
    \subfigure[\scriptsize{Pointmass Navigation}]
    {
        \includegraphics[width=0.18\columnwidth]{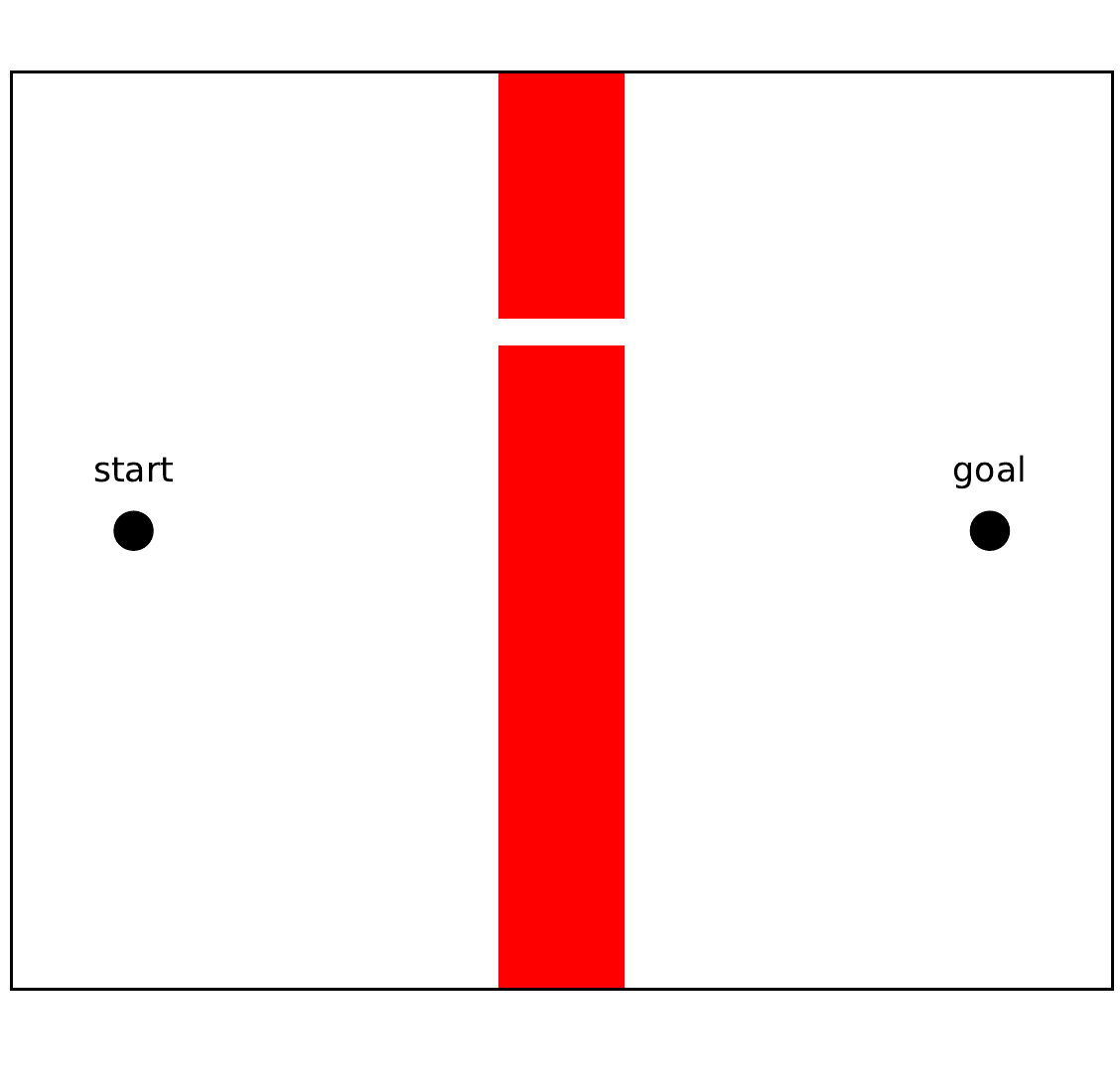}
        \label{fig:nav-env}
    }
    \subfigure[\scriptsize{Block Extraction}]
    {
        \includegraphics[width=0.18\columnwidth]{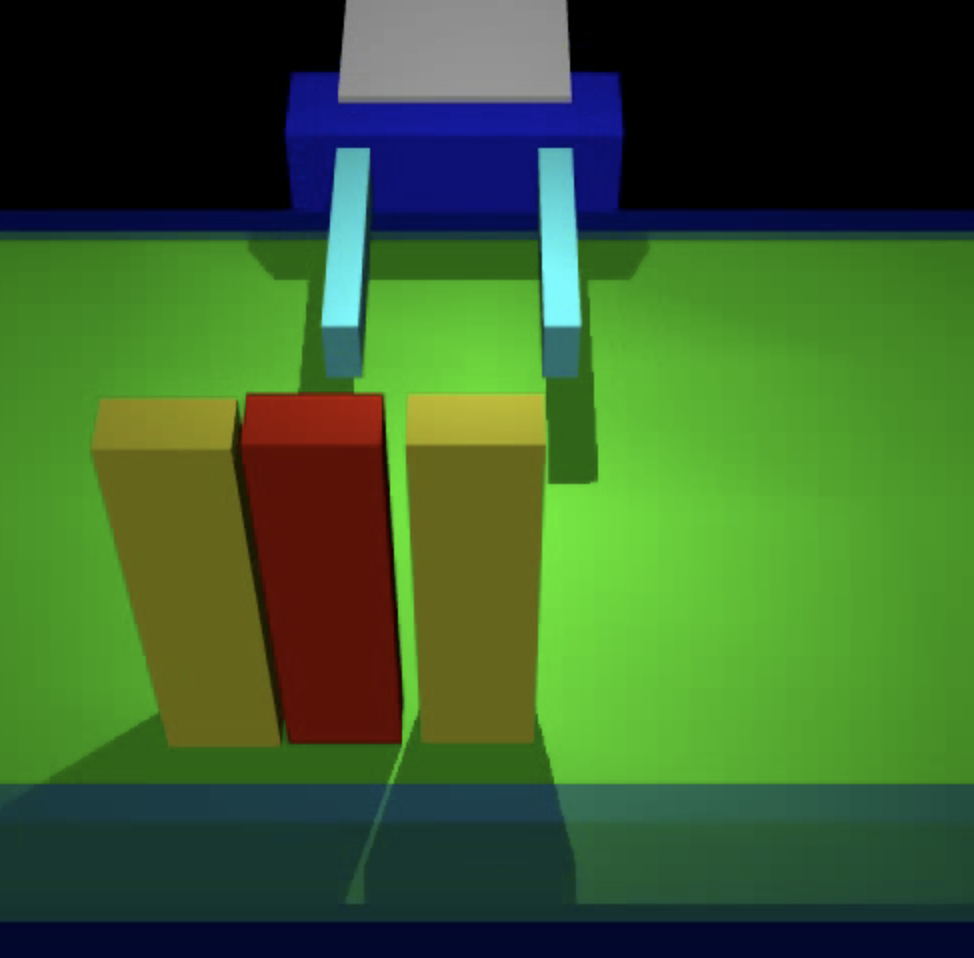}
        \label{fig:ext-env}
    }
    \subfigure[\scriptsize{Sequential Pushing}]
    {
        \includegraphics[width=0.18\columnwidth]{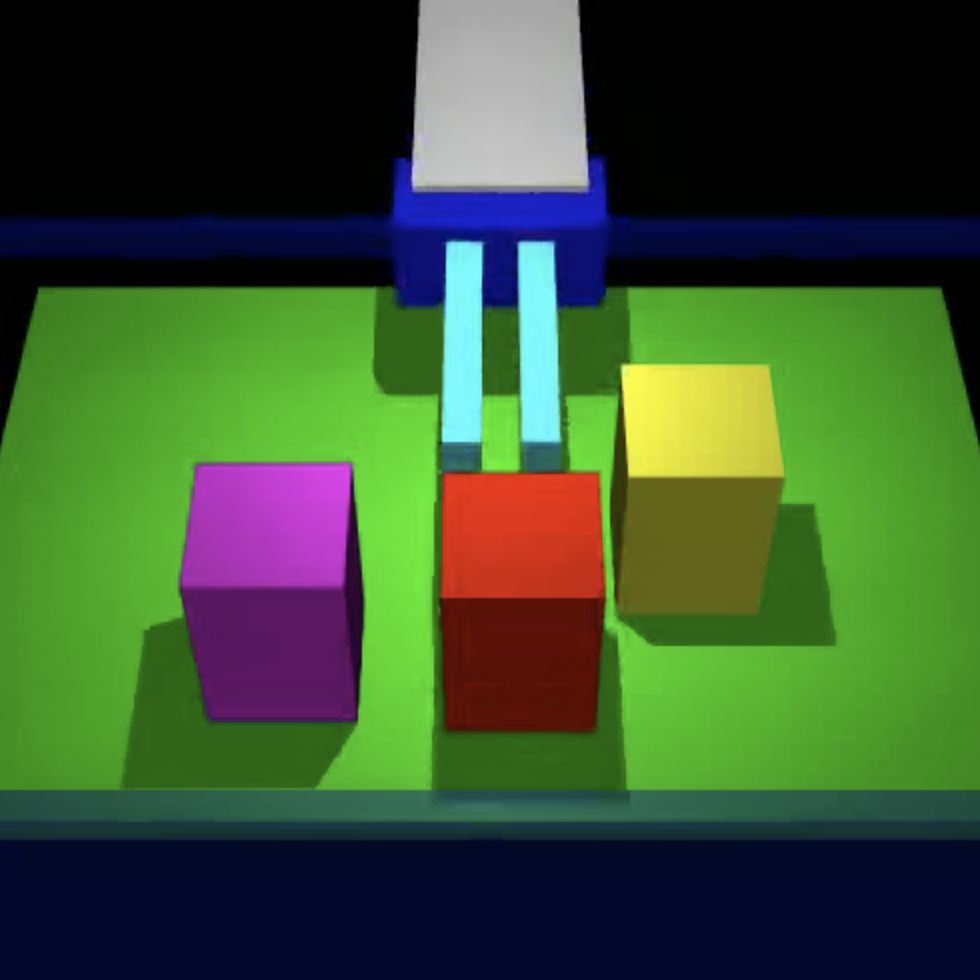}
        \label{fig:push-env}
    }
    \subfigure[\scriptsize{Door Opening}]
    {
        \includegraphics[width=0.18\columnwidth]{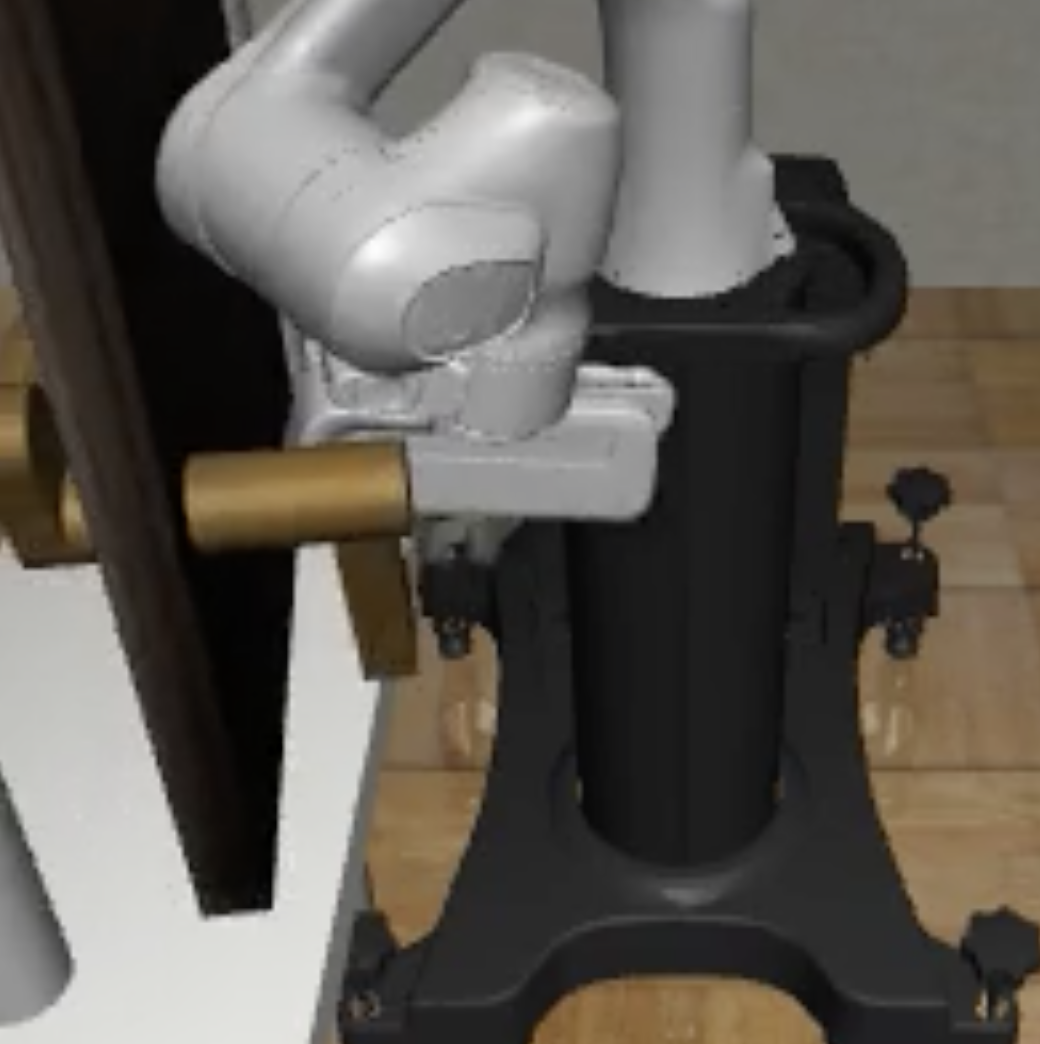}
        \label{fig:door-env}
    }
    \subfigure[\scriptsize{Block Lifting}]
    {
        \includegraphics[width=0.18\columnwidth]{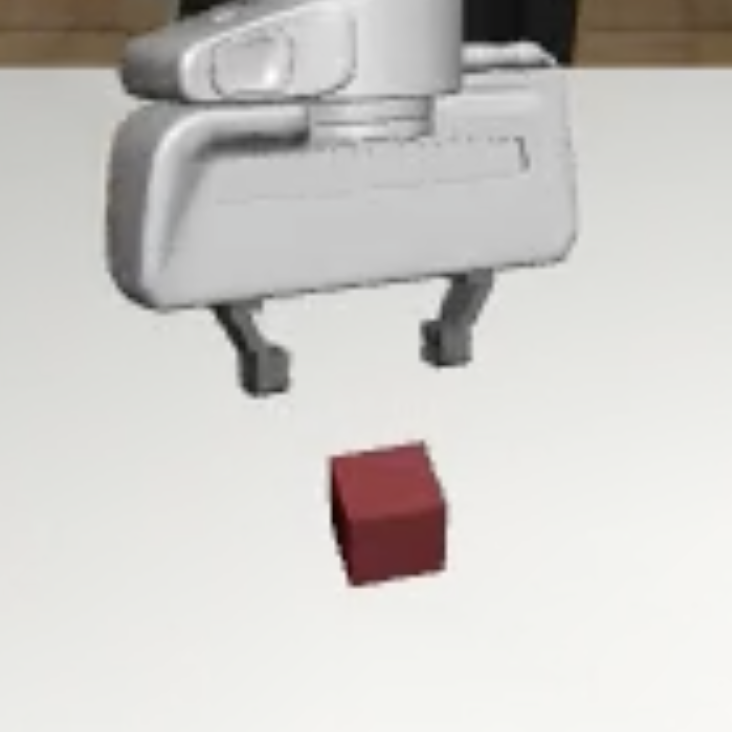}
        \label{fig:lift-env}
    }
    \caption{\algabbr{} Domains: We evaluate \algabbr{} on five continuous control domains: a pointmass navigation environment, and four high-dimensional robotic control domains. All domains are associated with relatively unshaped reward functions, which only indicate constraint violation, task completion, or completion of a subtask.}
    \label{fig:envs}
\end{figure}

We consider the five long-horizon continuous control tasks shown in Figure~\ref{fig:envs}. All tasks have relatively sparse rewards, making demonstrations critical for performance. We found that without demonstrations, SAC and TD3 made little to no progress on these tasks.

\paragraph{Pointmass Navigation: } The first domain is a pointmass 2D navigation task (Figure~\ref{fig:nav-env}) with time horizon $T=100$, where the objective is to navigate around the red barrier from start set $\mathcal{S}$ to a goal set $\mathcal{G}$ by executing 2D delta-position controls. If the agent collides with the barrier it receives a reward of $-100$ and the episode terminates. At each time step, the agent receives a reward of $-1$ if it is not in the goal set and $0$ if it is in the goal set. To increase the difficulty of the task, we perturb the state with zero-mean Gaussian noise at each timestep. The combination of noisy transitions and sparse reward signal makes this a very difficult exploration task where the agent must learn to make it through the slit without converging to the local optima of avoiding both the barrier and the slit.  

The demonstrator is implemented as a series of proportional controllers which guide it from the starting set to the slit, through the slit, and to the goal set. The actions are clipped to fit in the action space, and trajectories are nearly optimal. The agent is provided with $20$ demonstrations.


\paragraph{Object Manipulation in MuJoCo: }
We next consider two object manipulation tasks designed in the MuJoCo physics simulator \citep{mujoco}, where the objective is to extract a block from a tight configuration on a table (Block Extraction, Figure~\ref{fig:ext-env}) and push each of 3 blocks forward on the plane (Sequential Pushing, Figure~\ref{fig:push-env}). In the Block Extraction task, the action space consists of 3D delta position controls and an extra action dimension to control the degree to which the gripper is opened. In the Sequential Pushing environment, this extra action dimension is omitted and the gripper is kept closed. In the Block Extraction domain, the agent receives a reward of $-1$ for every timestep that it hasn't retrieved the red block and $0$ when it has. In the Sequential Pushing domain, the reward increases by 1 for each block the agent pushes forward. Thus, the agent receives a reward of $-3$ when it has made no progress and $0$ when it has completed the task. The Block Extraction task is adapted from~\citet{thananjeyan2021recovery} while the Sequential Pushing task is adapted from~\citet{LS3}. We use a time horizon of $T=50$ for the Block Extraction task and a longer $T=150$ for the Sequential Pushing task since it is of greater complexity.

The block extraction demonstrator is implemented as a series of proportional controllers guiding the arm to a position to grip the block, followed by an instruction to close the gripper and a controller to lift. We provide the agent with $50$ demonstrations with zero-mean noise injected in the controls to induce suboptimality. For the sequential pushing environment, the demonstrator uses a series of proportional controllers to slowly push one block forward, move backwards, line up with the next block, and repeat the motion until all blocks have been pushed. Because it moves slowly and moves far back from each block it pushes, demonstrations are very suboptimal. For Sequential Pushing, the agent is provided with $500$ demonstrations due to the increased difficulty of the task.

\paragraph{Robosuite Object Manipulation: } 
Finally, we consider two object manipulation tasks built on top of Robosuite~\citep{robosuite2020}, a collection of robot simulation tasks using the MuJoCo physics engine.
We consider the Door Opening task (Figure~\ref{fig:door-env}) and the Block Lifting task (Figure~\ref{fig:lift-env}). In the Door Opening task, a Panda robot with 7 DoF and a parallel-jaw gripper must turn the handle of a door in order to open it. The door's location is randomized at the start of each episode. The agent receives a reward of -1 if it has not opened the door and a reward of 0 if it has. In the Block Lifting task, the same Panda robot is placed in front of a table with a single block on its surface. The robot must pick up the block and lift it above a certain threshold height. The block's location is randomized for each episode and the agent receives a reward of $-1$ for every timestep it has not lifted the block and a reward of $0$ when it has. Both Robosuite tasks use a time horizon of $T=50$.

For both Robosuite tasks, demonstrators are trained using SAC on a version of the task with a hand-designed dense reward function, as in the Robosuite benchmarking experiments~\citep{robosuite2020}. In order to ensure suboptimality, we stop training the demonstrator policy before convergence. For each Robosuite environment we use the trained demonstrator policies to generate 100 suboptimal demonstrations for training MCAC and the baselines.

\begin{figure}[b!]
    \centering
    \includegraphics[width=\columnwidth]{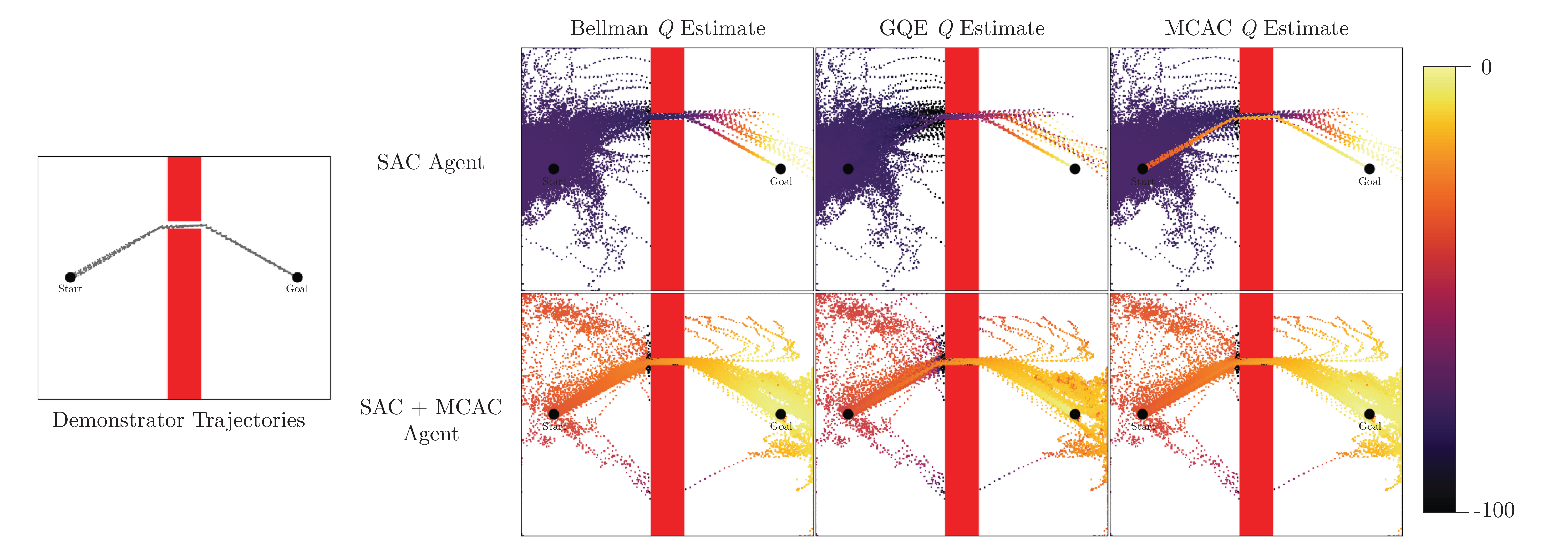}
    \caption{\textbf{\algabbr{} Replay Buffer Visualization: }Scatter plots showing Bellman, GQE and MCAC $Q$ estimates on the entire replay buffer, including offline demonstrations, for SAC learners with and without the \algabbr{} modification after 50000 timesteps of training. The top row shows data and $Q$ estimates obtained while training a baseline SAC agent without \algabbr{}, while the bottom row shows the same when SAC is trained with \algabbr{}. The left column shows Bellman $Q$ estimates on each replay buffer sample while the middle column shows GQE estimates and the right column shows \algabbr{} estimates. Results suggest that \algabbr{} is helpful for propagating rewards along demonstrator trajectories.}
    \label{fig:replay_buffer}
\end{figure}

\subsection{Algorithm Comparisons}
\label{subsec:comparisons}

\begin{figure*}[t]
\centering
    \subfigure[Pointmass Navigation]
    {
        \includegraphics[width=.31\columnwidth]{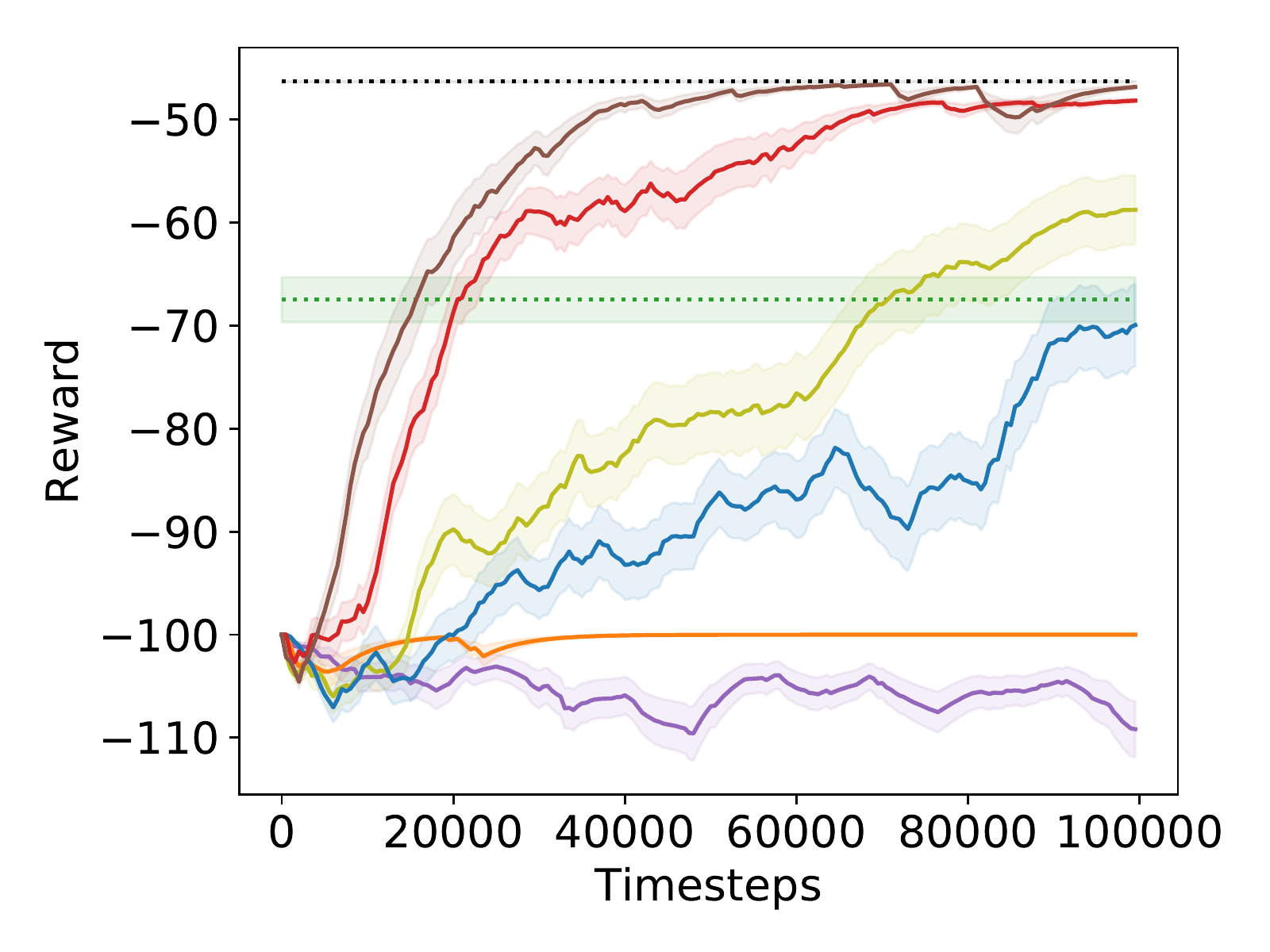}
        \label{fig:nav-results}
    }
    \subfigure[Block Extraction]
    {
        \includegraphics[width=.31\columnwidth]{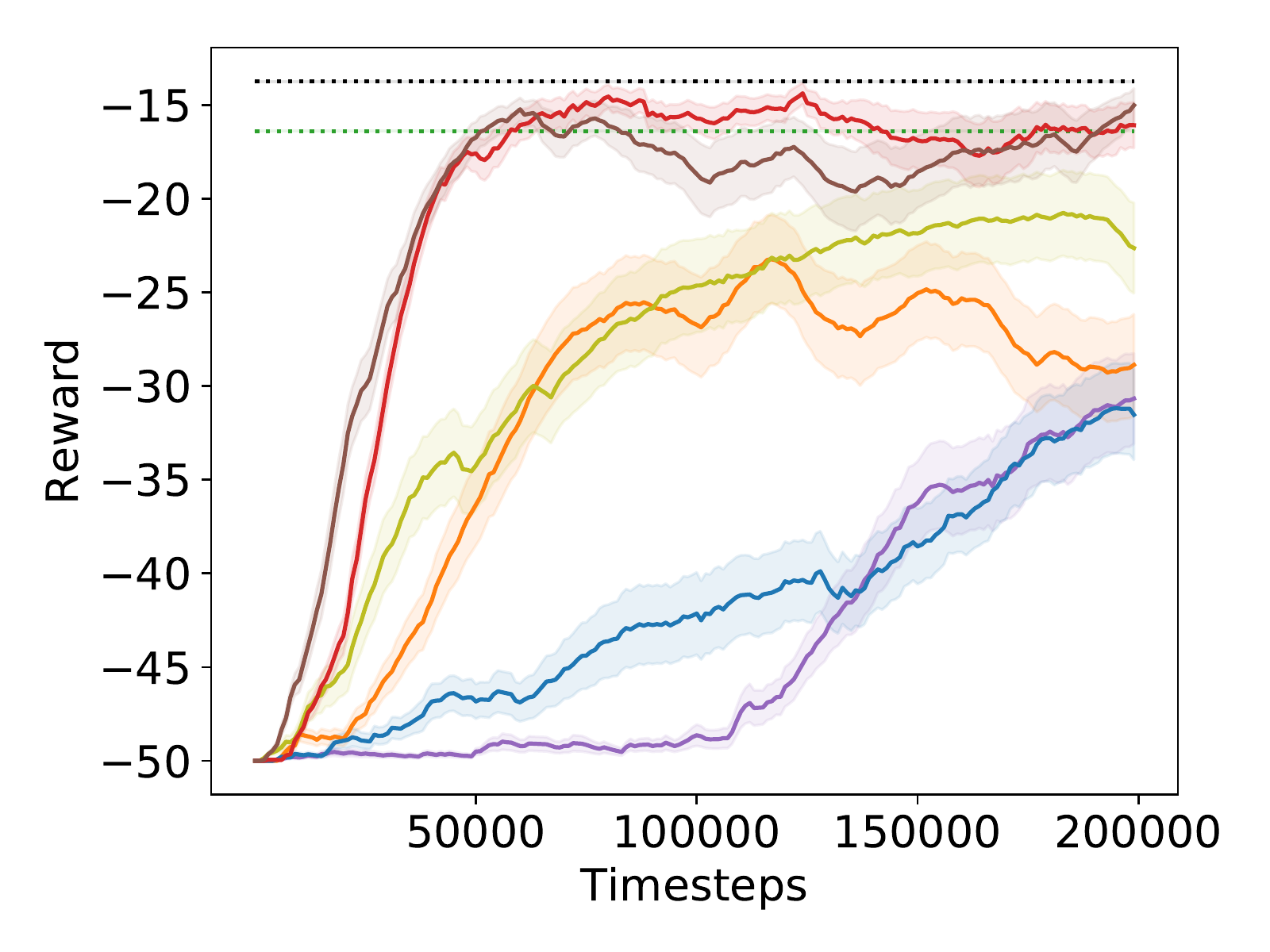}
        \label{fig:ext-result}
    }
    \subfigure[Sequential Pushing]
    {
        \includegraphics[width=.31\columnwidth]{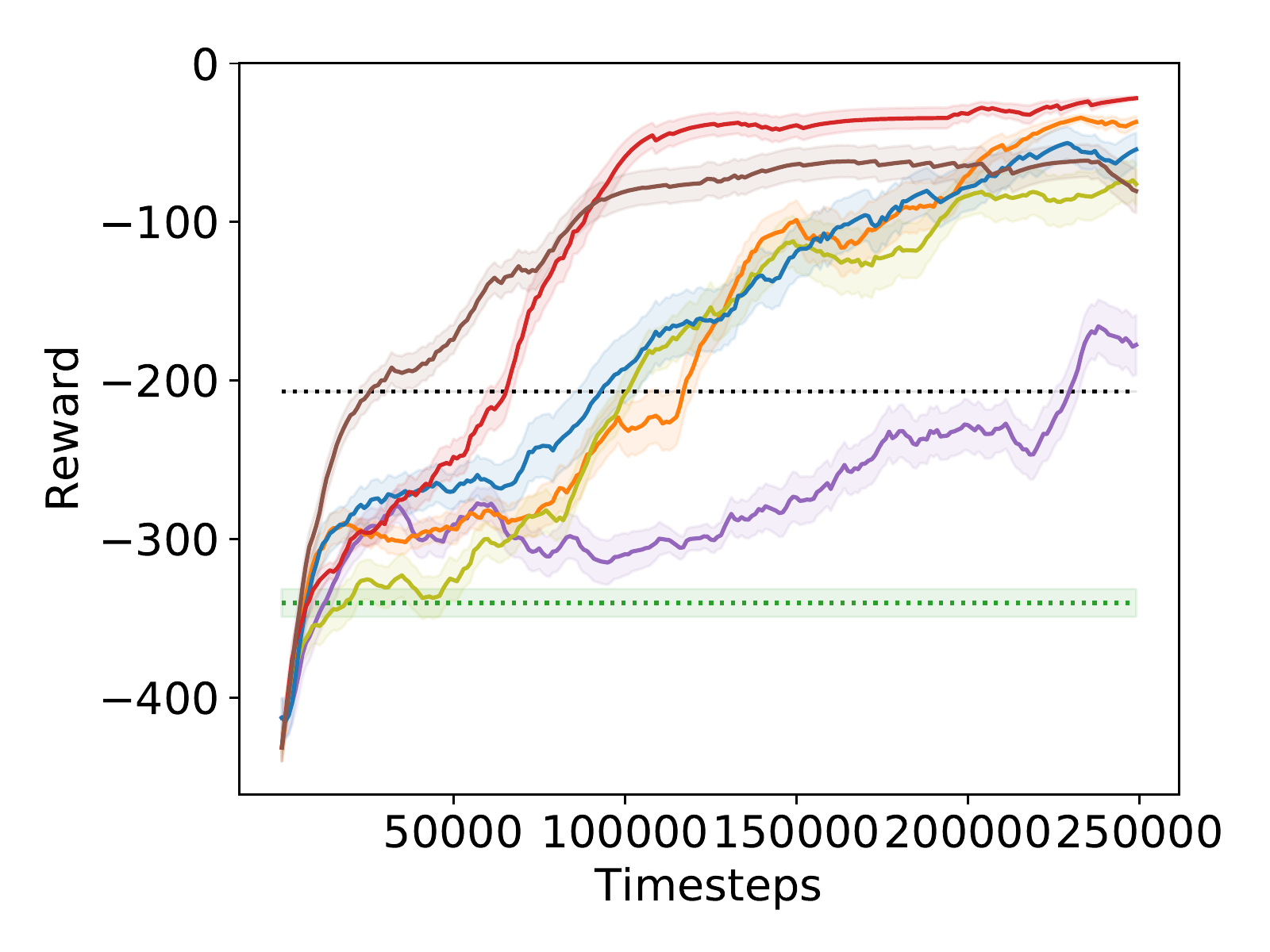}
        \label{fig:push-results}
    }
    \subfigure[Door Opening]
    {
        \includegraphics[width=.31\columnwidth]{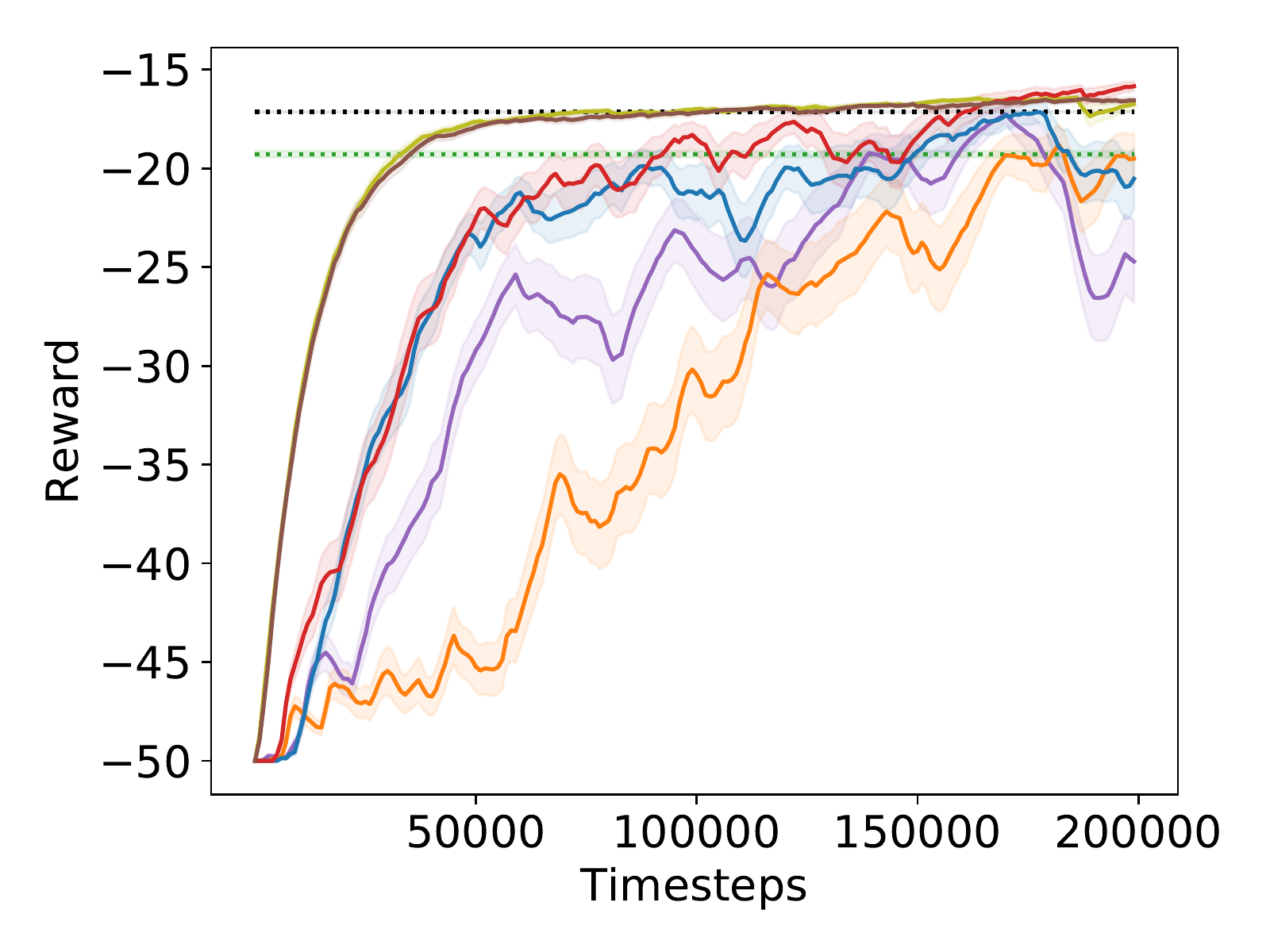}
        \label{fig:door-results}
    }
    \subfigure[Block Lifting]
    {
        \includegraphics[width=.31\columnwidth]{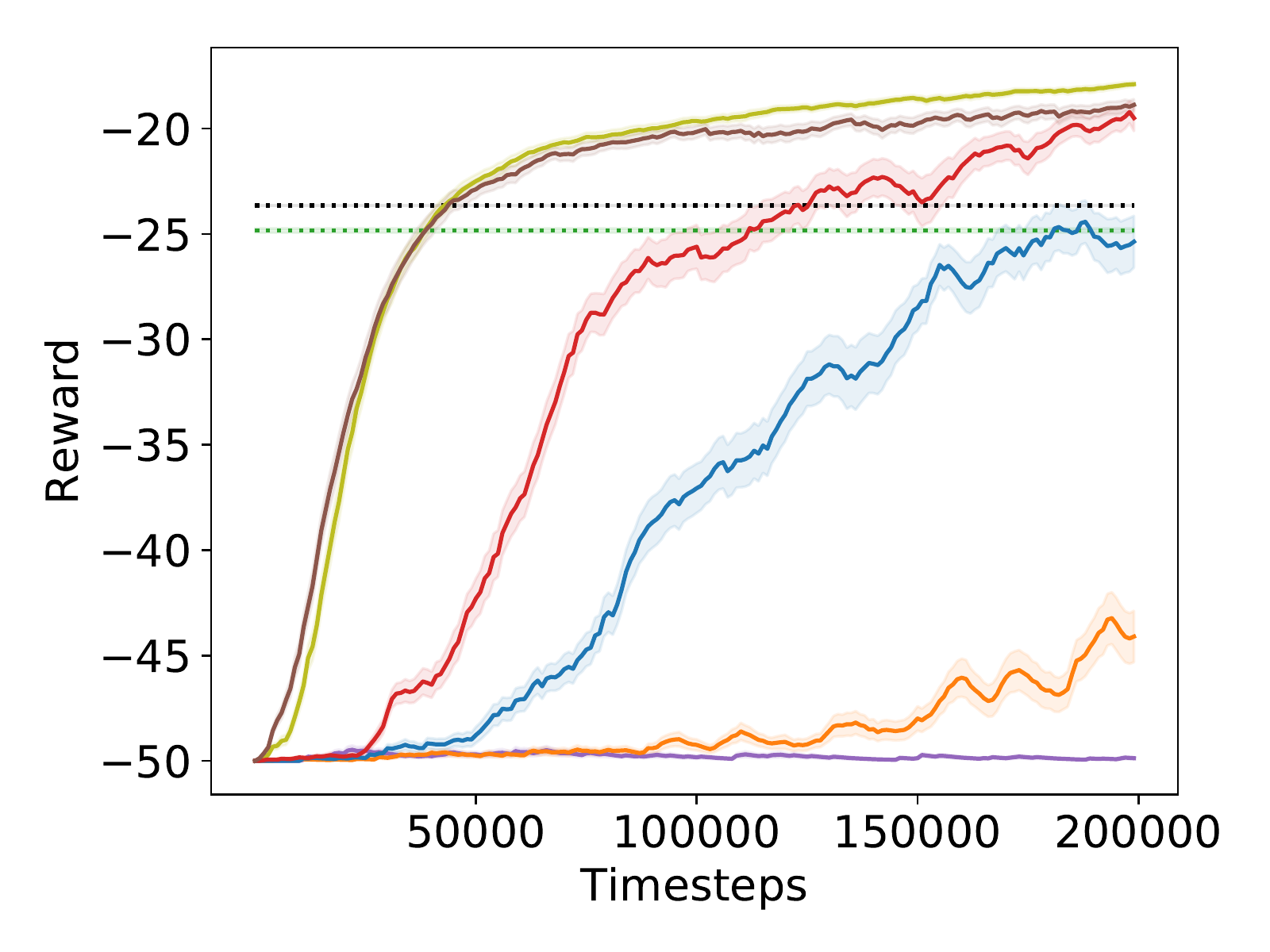}
        \label{fig:lift-results}
    }
    \includegraphics[width=\columnwidth]{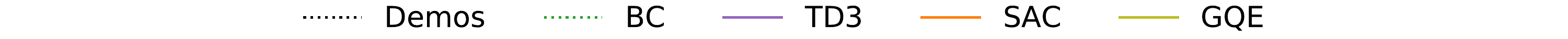}
    \includegraphics[width=\columnwidth]{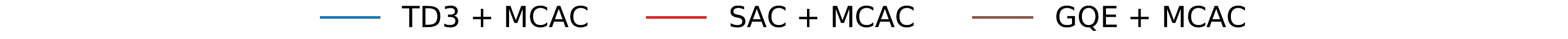}
     \caption{\textbf{\algabbr{} and Standard RL Algorithms Results: }Learning curves showing the exponentially smoothed (smoothing factor $\gamma=0.9$) mean and standard error across 10 random seeds. We find that \algabbr{} improves the learning efficiency of TD3, SAC, and GQE across all 5 environments.}
    \label{fig:commonrl-results}
\end{figure*}

We empirically evaluate the following baselines both individually and in combination with \algabbr{}. All methods are provided with the same demonstrations which are collected as described in Section~\ref{subsec:domains}. See the supplement for more in depth details on implementation and training. 

\textbf{Behavior Cloning: }  Direct supervised learning on the offline suboptimal demonstrations.

\textbf{Twin Delayed Deep Deterministic Policy Gradients (TD3)~\citep{fujimoto2018addressing}: }State of the art actor-critic algorithm which trains a deterministic policy to maximize a learned critic.

\textbf{Soft Actor-Critic (SAC)~\citep{haarnoja2017soft}: } State of the art actor-critic algorithm which trains a stochastic policy which maximizes a combination of the $Q$ value of the policy and the expected entropy of the policy to encourage exploration.

\textbf{Generalized $Q$ Estimation (GQE): } A complex return estimation method from~\citet{GAE} for actor-critic methods, which computes a weighted average over TD-$i$ estimates for a range of $i$. GQE is implemented on top of SAC (see supplement for more details). We tune the range of horizons considered and the eligibility trace value (corresponding to $\lambda$ in GAE).

\textbf{Overcoming Exploration from Demonstrations (OEFD)~\citep{nair2018overcoming}: }OEFD  builds on DDPG~\citep{DDPG} by adding a loss which encourages imitating demonstrations and a learned filter which determines when to activate this loss. Our implementation does not include hindsight experience replay since it is not applicable to most of our environments.

\textbf{Conservative $Q$ Learning (CQL)~\citep{CQL}: }A recent offline RL algorithm that addresses value overestimation with a conservative $Q$ function. Here we also update CQL online after pre-training offline on demonstrations in order to provide a fair comparison with other algorithms.

\textbf{Advantage Weighted Actor-Critic (AWAC)~\citep{AWAC}: }A recent offline reinforcement learning algorithm designed for fast online fine-tuning. 

We also implement versions of each of the above RL algorithms with \algabbr{} (TD3 + \algabbr{}, SAC + \algabbr{}, GQE + \algabbr{}, OEFD + \algabbr{}, CQL + \algabbr{}, AWAC + \algabbr{}).

The behavior cloning comparison serves to determine whether online learning is beneficial in general, while the other comparisons study whether \algabbr{} can be used to accelerate reinforcement learning for commonly used actor-critic algorithms (TD3, SAC, and GQE, which is essentially SAC with complex returns) and  for recent algorithms for RL from demonstrations (OEFD, CQL and AWAC).

\subsection{Results}
\label{subsec:results}
In Section~\ref{subsubsec:didactic}, we study \algabbr{} on a simple didactic environment to better understand how it affects the learned Q-values. We then study whether \algabbr{} can be used to accelerate exploration on a number of continuous control domains. In Section~\ref{subsubsec:commonrl}, we study whether \algabbr{} enables more efficient learning when built on top of widely used actor-critic RL algorithms (SAC, TD3, and GQE). Then in Section~\ref{subsubsec:demorl}, we study whether \algabbr{} can provide similar benefits when applied to recent RL from demonstration algorithms (OEFD, CQL, and AWAC). Additionally, in the supplement we provide experiments involving other baselines, investigating the sensitivity of \algabbr{} to the quality and quantity of demonstration data, and investigating its sensitivity to pretraining.

\subsubsection{\algabbr{} Didactic Example}
\label{subsubsec:didactic}
In order to better understand the way \algabbr{} affects $Q$ estimates, we visualize $Q$ estimates when \algabbr{} is applied to SAC after 50000 timesteps of training in the Pointmass Navigation environment, in Figure~\ref{fig:replay_buffer}. Here we visualize $Q$-values for the entire replay buffer, including offline demonstrations,

When training without \algabbr{} (top row), the agent is unable to learn a useful $Q$ function and thus does not learn to complete the task (the only successful trajectories shown are offline data). However, even when this is the case, the \algabbr{} estimate is able to effectively propagate reward signal backwards along the demonstrator trajectories, predicting higher rewards early on (top right). We see that the GQE estimates (top middle) are somewhat more effective than the Bellman ones at propagating reward, but not as effective as \algabbr{}. When the agent is trained with \algabbr{} (bottom row), the agent learns a useful $Q$ function that it uses to reliably complete the task (bottom left). As expected with a high-performing policy, its Bellman estimates, GQE estimates and \algabbr{} estimates are similar.

\subsubsection{\algabbr{} and Standard RL Algorithms}
\label{subsubsec:commonrl}
In Figure~\ref{fig:commonrl-results}, we study the impact of augmenting SAC, TD3 and GQE with the \algabbr{} target $Q$-function. Note that all methods, both with and without \algabbr{}, we initialize their replay buffers with the same set of demonstrations. Results suggest that \algabbr{} is able to accelerate learning for both TD3 and SAC across all environments, and is able to converge to performance either on-par with or better than the demonstrations. In the Pointmass Navigation and Block Lifting tasks, SAC and TD3 make no task progress without \algabbr{}. \algabbr{} also accelerates learning for GQE for the Pointmass Navigation, Block Extraction, and Sequential Pushing environments. In the Door Opening and Block Lifting environments, \algabbr{} leaves performance largely unchanged since GQE already achieves performance on par with the next best algorithm without \algabbr{}.


\subsubsection{\algabbr{} and RL From Demonstrations Algorithms}
\label{subsubsec:demorl}

\begin{figure*}[t!]
\centering
    \subfigure[Pointmass Navigation]
    {
        \includegraphics[width=.31\columnwidth]{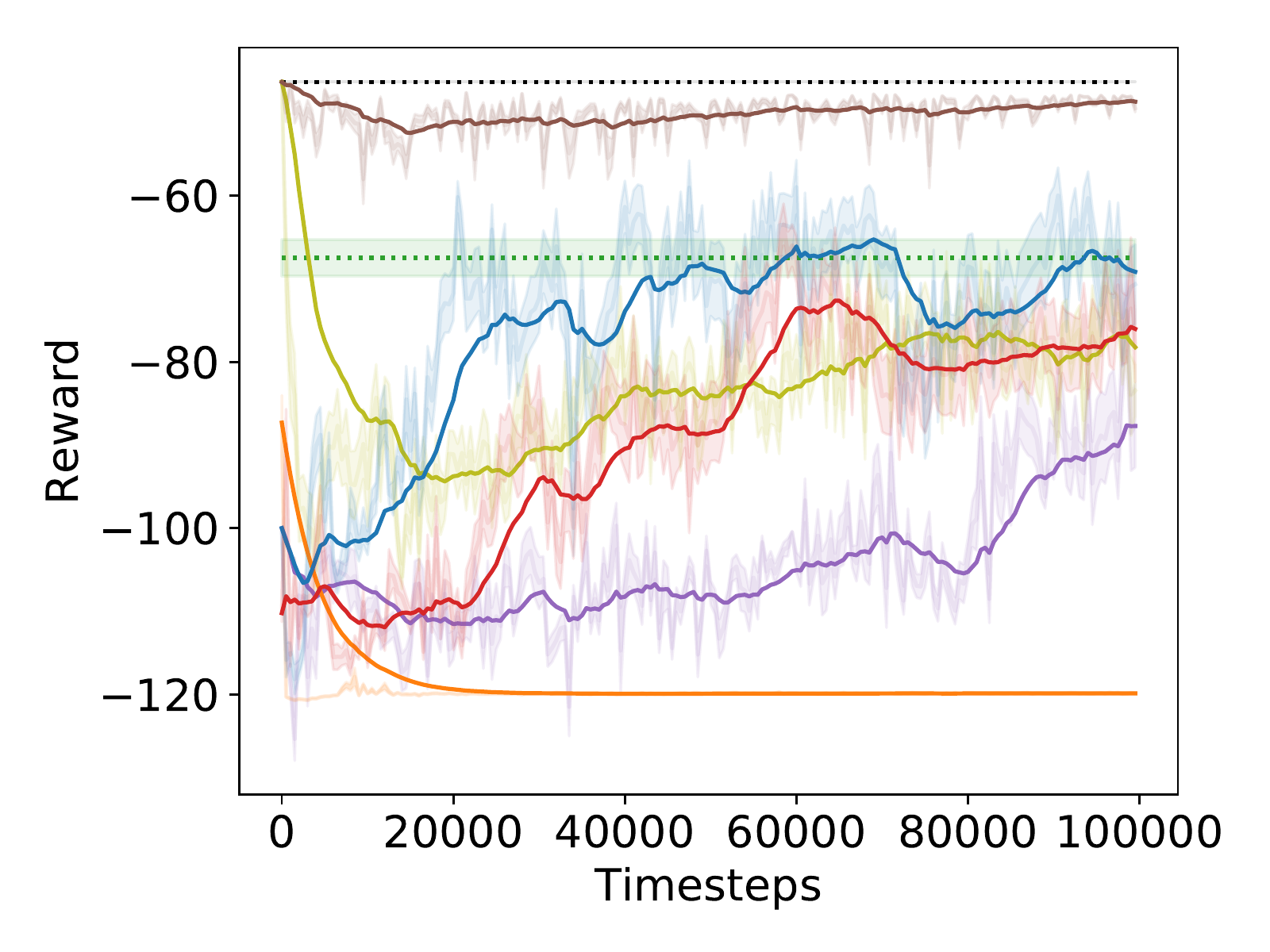}
        \label{fig:nav-results}
    }
    \subfigure[Block Extraction]
    {
        \includegraphics[width=.31\columnwidth]{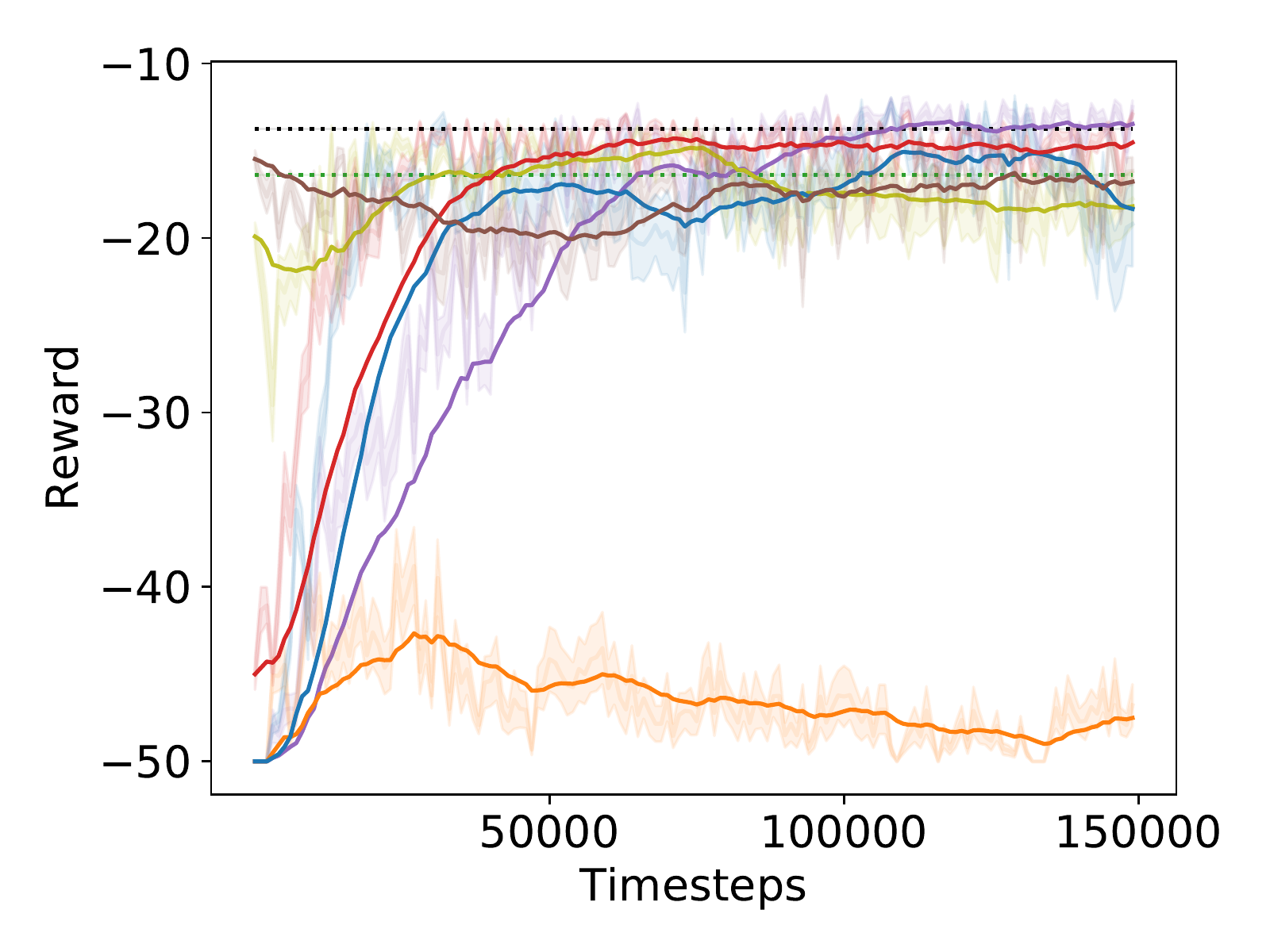}
        \label{fig:ext-result}
    }
    \subfigure[Sequential Pushing]
    {
        \includegraphics[width=.31\columnwidth]{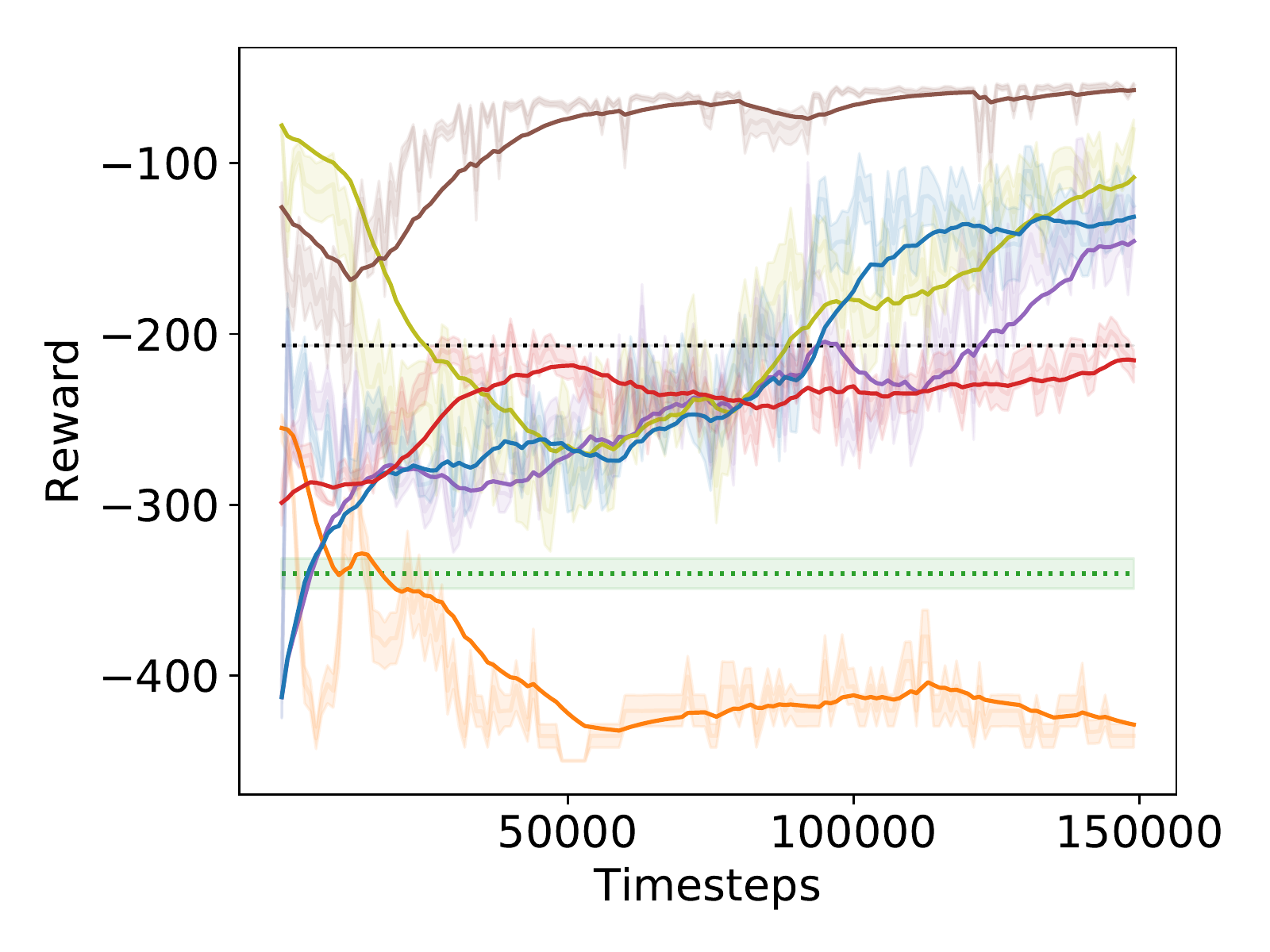}
        \label{fig:push-results}
    }
    \subfigure[Door Opening]
    {
        \includegraphics[width=.31\columnwidth]{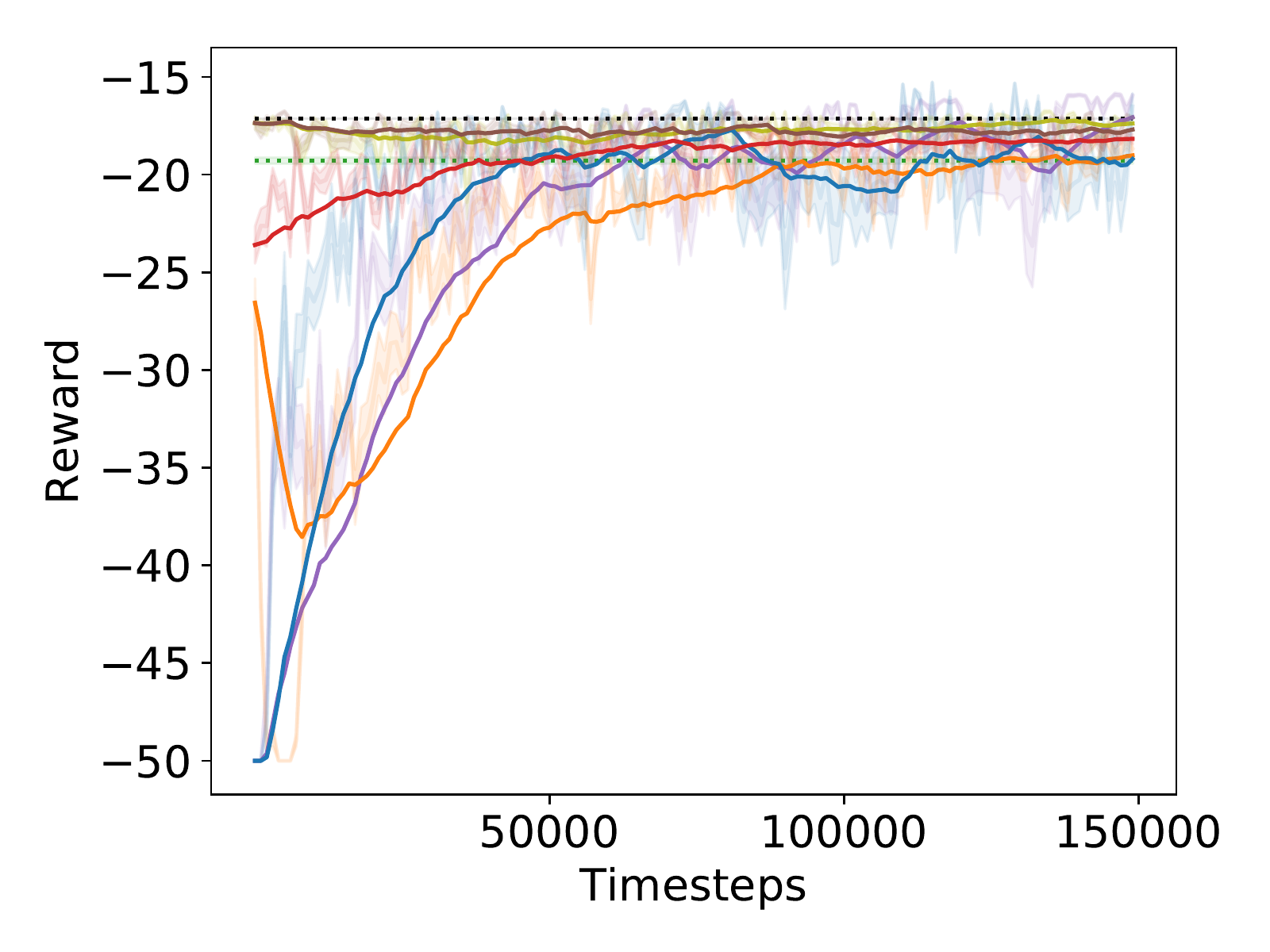}
        \label{fig:door-results}
    }
    \subfigure[Block Lifting]
    {
        \includegraphics[width=.31\columnwidth]{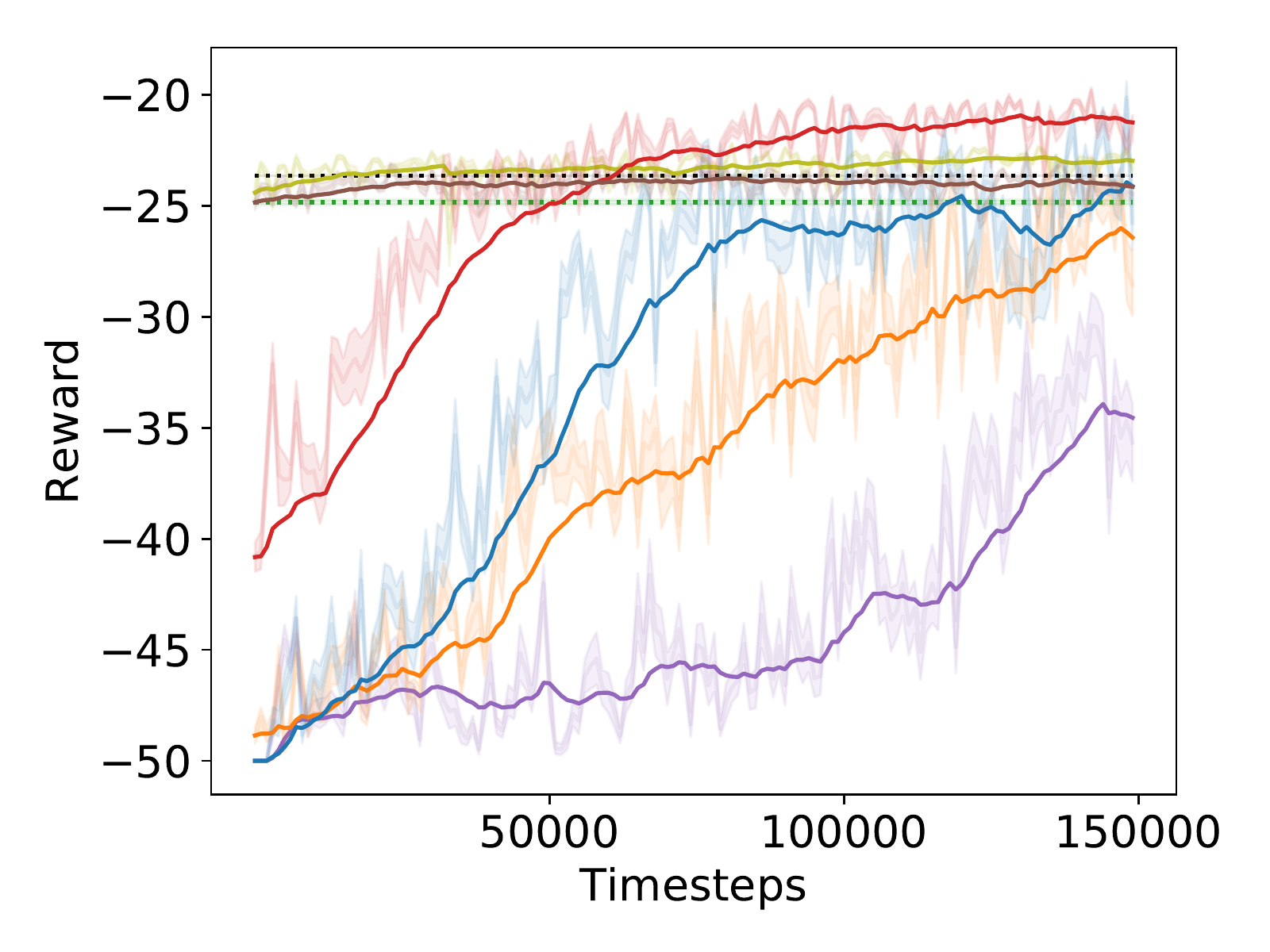}
        \label{fig:lift-results}
    }
    \includegraphics[width=\columnwidth]{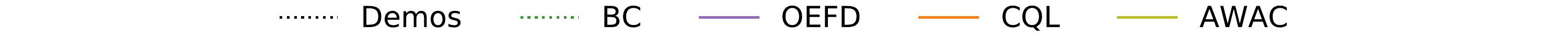}
    \includegraphics[width=\columnwidth]{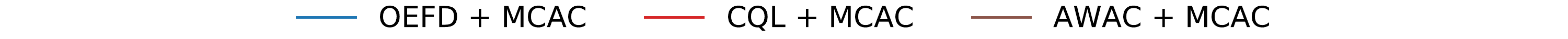}
     \caption{\textbf{\algabbr{} and RL from Demonstrations Algorithm Results: }Learning curves showing the exponentially smoothed (smoothing factor $\gamma=0.9$) mean and standard error across 10 random seeds. When OEFD or AWAC achieve high performance almost immediately, \algabbr{} has little impact on performance. However, when OEFD and AWAC are unable to learn efficiently, \algabbr{} accelerates and stabilizes policy learning.}
    \label{fig:demorl-results}
\end{figure*}

In Figure~\ref{fig:demorl-results}, we study the impact of augmenting OEFD~\citep{nair2018overcoming}, CQL~\citep{CQL} and AWAC~\citep{AWAC} with the \algabbr{} target $Q$-function. Results suggest that \algabbr{} improves the learning efficiency of OEFD on the Pointmass Navigation, Sequential Pushing, and Block Lifting tasks, but does not have a significant positive or negative affect on performance for the Block Extraction and Door Opening tasks. \algabbr{} improves the performance of AWAC on the Pointmass Navigation and Sequential Pushing environments, stabilizing learning while the versions without \algabbr{} see performance fall off during online fine tuning. On the other 3 environments where AWAC is able to immediately converge to a stable policy after offline pre-training, \algabbr{} has no significant negative effect on its performance. In all tasks, \algabbr{} improves the performance of CQL. In particular, for the Pointmass Navigation, Block Extraction and Sequential Pushing tasks, CQL makes almost no progress while the version with \algabbr{} learns to complete the task reliably.

\section{Discussion, Limitations, and Ethical Considerations}
\label{sec:discussion}
We present \algname{} (\algabbr{}), a simple, yet highly effective, change that can be applied to any actor-critic algorithm in order to accelerate reinforcement learning from demonstrations for sparse reward tasks. We present empirical results suggesting that \algabbr{} often significantly improves performance when applied to three state-of-the-art actor-critic RL algorithms and three RL from demonstrations algorithms on five different continuous control domains.

Despite strong empirical performance, \algabbr{} also has some limitations. We found that while encouraging higher $Q$-value estimates is beneficial for sparse reward tasks, when rewards are dense and richly informative, \algabbr{} is not helpful and can even hinder learning by overestimating $Q$-values. From an ethical standpoint, reinforcement learning algorithms such as \algabbr{} can be used to automate aspects of a variety of interactive systems, such as robots, online retail systems, or social media platforms. However, poor performance when policies haven't yet converged, or when hand-designed reward functions do not align with true human intentions, can cause significant harm to human users. \new{Lastly, while \algabbr{} shows convincing empirical performance, in future work it would be interesting to provide theoretical analysis on its convergence properties and on how \algabbr{} can be used to learn $Q$ functions which better navigate the bias-variance tradeoff.}
\section*{Acknowledgements}

We would like to thank all reviewers for their invaluable feedback, which helped us significantly strengthen the paper during the author response period. This research was performed at the AUTOLAB at UC Berkeley in affiliation with the Berkeley AI Research (BAIR) Lab, and the CITRIS "People and Robots" (CPAR) Initiative. The authors were supported in part by donations from Google, Toyota Research Institute, and NVIDIA.

\bibliographystyle{plainnat}
\bibliography{main}


\onecolumn
\newpage
\appendix

\section{Implementation Details}\label{app:implementation_details}

For all algorithms, all $Q$ functions and policies are approximated using deep neural networks with 2 hidden layers of size 256. They are all updated using the Adam optimizer from \cite{Kingma2015AdamAM}.

\subsection{Behavioral Cloning}
We used a straightforward implementation of behavioral cloning, regressing with a mean square error loss. For all experiments learners were provided with the same number of demonstrators as the other algorithms and optimized for $10000$ gradient steps using a learning rate of $1\times10^{-4}$.

\subsection{Twin Delayed Deep Deterministic Policy Gradients} \label{sec:app-td3}

We use the author implementation of TD3 from \cite{fujimoto2018addressing}, modifying it to implement \algabbr{}. In order to maintain the assumption about complete trajectories described in the main text, we modify the algorithm to only add to the replay buffer at the end of sampled trajectories, but continue to update after each timestep. We found the default hyperparameters from the repository to be sufficient in all environments.  

\subsection{Soft Actor Critic} \label{sec:app-sac}
For all SAC experiments we used a modified version of the SAC implementation from \citet{sac_github} which implements SAC with a double-$Q$ critic update function to combat $Q$ overestimation. Additionally, we modify the algorithm to satisfy the trajectory assumption as in Section~\ref{sec:app-td3}. We mostly use the default hyperparameters from \cite{haarnoja2017soft}, but tuned $\alpha$ and $\tau$. Parameter choices are shown in Table~\ref{tab:hyperparams_sac}.

\begin{table}[h]
    \centering
    \caption{Hyperparameters for SAC}
    \begin{tabular}{c c c c}
        \hline
        Parameter & Navigation & MuJoCo & Robosuite \\
        \hline
        \hline
        Learning Rate & $3\times 10^{-4}$ & $3\times 10^{-4}$ & $3\times 10^{-4}$ \\
        Automatic Entropy Tuning & False & False & False \\
        Batch Size & $256$ & $256$ & $256$ \\
        Hidden Layer Size & $256$ & $256$ & $256$ \\
        \# Hidden Layers & $2$ & $2$ & $2$ \\
        \# Updates Per Timestep & $1$ & $1$ & $1$ \\
        $\alpha$ & $0.2$ & $0.1$ & $0.05$ \\
        $\gamma$ & $0.99$ & $0.99$ & $0.99$ \\
        $\tau$ & $5 \times 10^{-2}$ & $5\times 10^{-3}$ & $5\times 10^{-2}$ \\
        \hline
    \end{tabular}
    \label{tab:hyperparams_sac}
\end{table}

\subsection{Generalized $Q$ Estimation}\label{app:gqe}

Similarly to the advantage estimation method in \citet{GAE}, we estimate $Q$ values by computing a weighted average over a number of $Q$ estimates estimated with $k$-step look-aheads. Concretely, if a $Q_t^{(k)}$ is a $Q$ estimate with a $k$-step look-ahead given by
\begin{equation}
    Q_t^{(k)} = \sum^{k-1}_{i=0} r_{t+i} + \gamma^{k}Q_\theta(s_{t+k}, a_{t+k}),
\end{equation}
we compute the $n$-step GQE estimate $Q^{\mathrm{GQE}}_t$ as 
\begin{equation}
    Q_t^{\mathrm{GQE}} = \frac{1-\lambda}{1-\lambda^n}\sum_{k=1}^n \lambda^{k-1} Q_t^{(k)}.
\end{equation}
We built GQE on top of SAC, using the SAC $Q$ estimates for the values of $Q_\theta$. However, in principle this method can be applied to other actor-critic methods.

Where applicable we used the hyperparameters from SAC, and tuned the values of $\lambda$ and $n$ as hyperparameters, trying values in the sets $\lambda \in \{0.8, 0.9, 0.95, 0.99, 0.999\}$ and $n$ values in the set $n \in \{8, 16, 32, 64\}$. The chosen parameters for each environment are given in Table~\ref{tab:hyperparams_gqe}.

\begin{table}[h]
    \centering
    \caption{Hyperparameters for GQE}
    \begin{tabular}{cccccc}
        \hline
        Parameter & Navigation & Extraction & Push & Door & Lift \\
        \hline
        \hline
        $\lambda$ & 0.9 & 0.95 & 0.95 & 0.9 & 0.9\\
        $n$ & 32 & 8 & 16 & 16 & 16\\
       
        \hline
    \end{tabular}
    \label{tab:hyperparams_gqe}
\end{table}


\subsection{Overcoming Exploration with Demonstrations}\label{}

We implement the algorithm from \cite{nair2018overcoming} on top of the implementation of TD3 described in Section~\ref{sec:app-td3}. Because it would provide an unfair advantage over comparisons, the agent is not given the ability to reset to arbitrary states in the replay buffer. Since our setting is not goal-conditioned, our implementation does not include hindsight experience replay. For the value $\lambda$ balancing the actor critic policy update loss and behavioral cloning loss, we use $\lambda=1.0$. \new{In all experiments the agent is pretrained on offline data for $10000$ gradient steps.}

\subsection{Conservative $Q$-learning} Offline reinforcement learning algorithm that produces a lower bound on the value of the current policy. We used the implementation from \citep{SAC_CQL}, which implements CQL on top of SAC as is done in the original paper, modified for additional online-finetuning. We used the default hyperparameters from~\citep{CQL} in all environments and pretrained the agent on offline data for 10000 gradient steps.

\subsection{Advantage Weighted Actor Critic}
\label{subsec:awac}

For AWAC experiments we use the implementation from \cite{Sikchi_pytorchAWAC}, once again modifying it to maintain the assumption about complete trajectories and to implement \algabbr{}. We found the default hyperparameter values to be sufficient in all settings. \new{In all experiments the agent is pretrained on offline data for $10000$ gradient steps.}


\section{Additional Experiments}

\subsection{No Demonstrations}
\label{subsec:no-demos}

\begin{figure*}[htb!]
\centering
    \subfigure[Pointmass Navigation]
    {
        \includegraphics[width=.3\columnwidth]{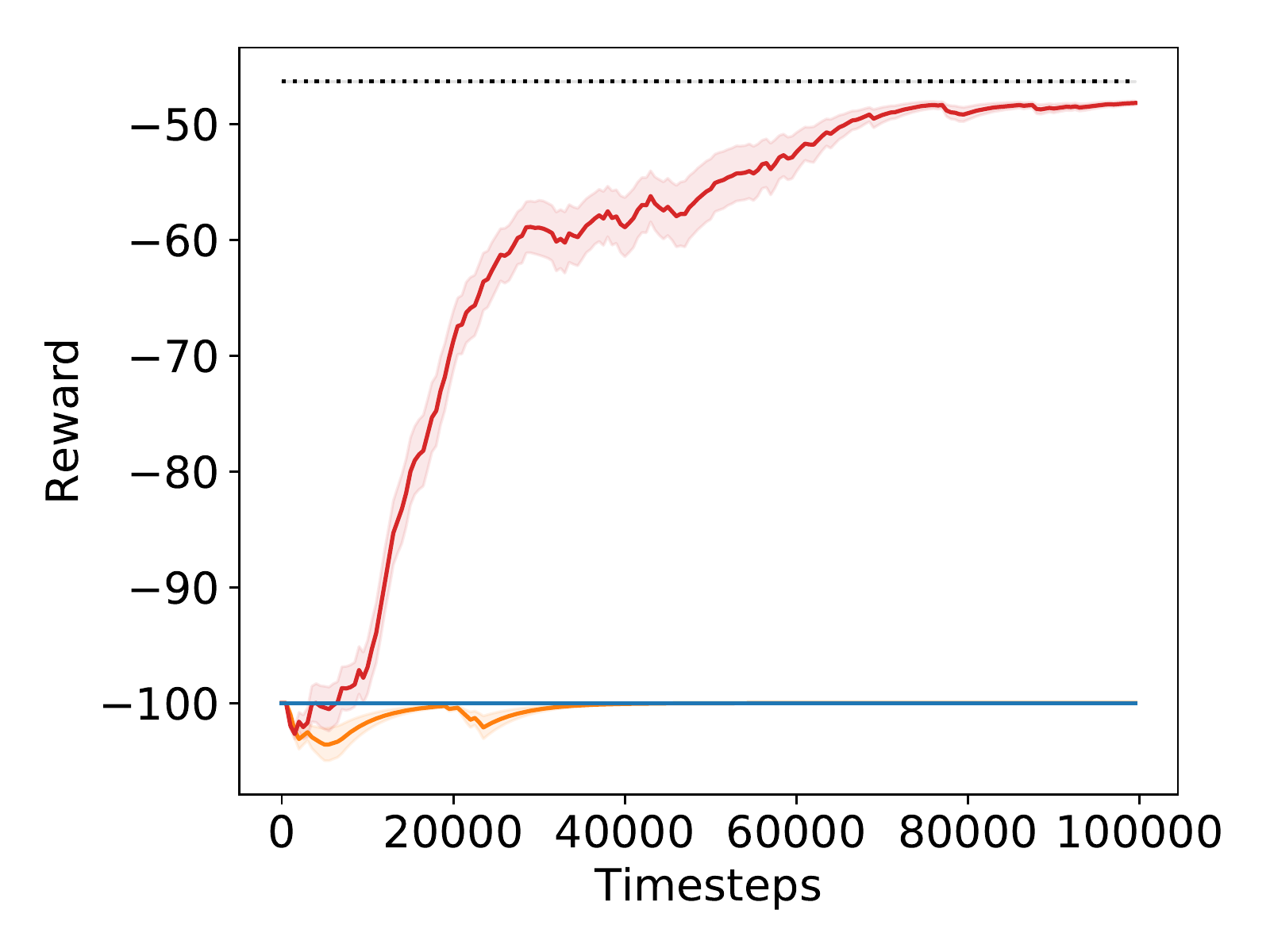}
        \label{fig:nav-results-nd}
    }
    \subfigure[Block Extraction]
    {
        \includegraphics[width=.3\columnwidth]{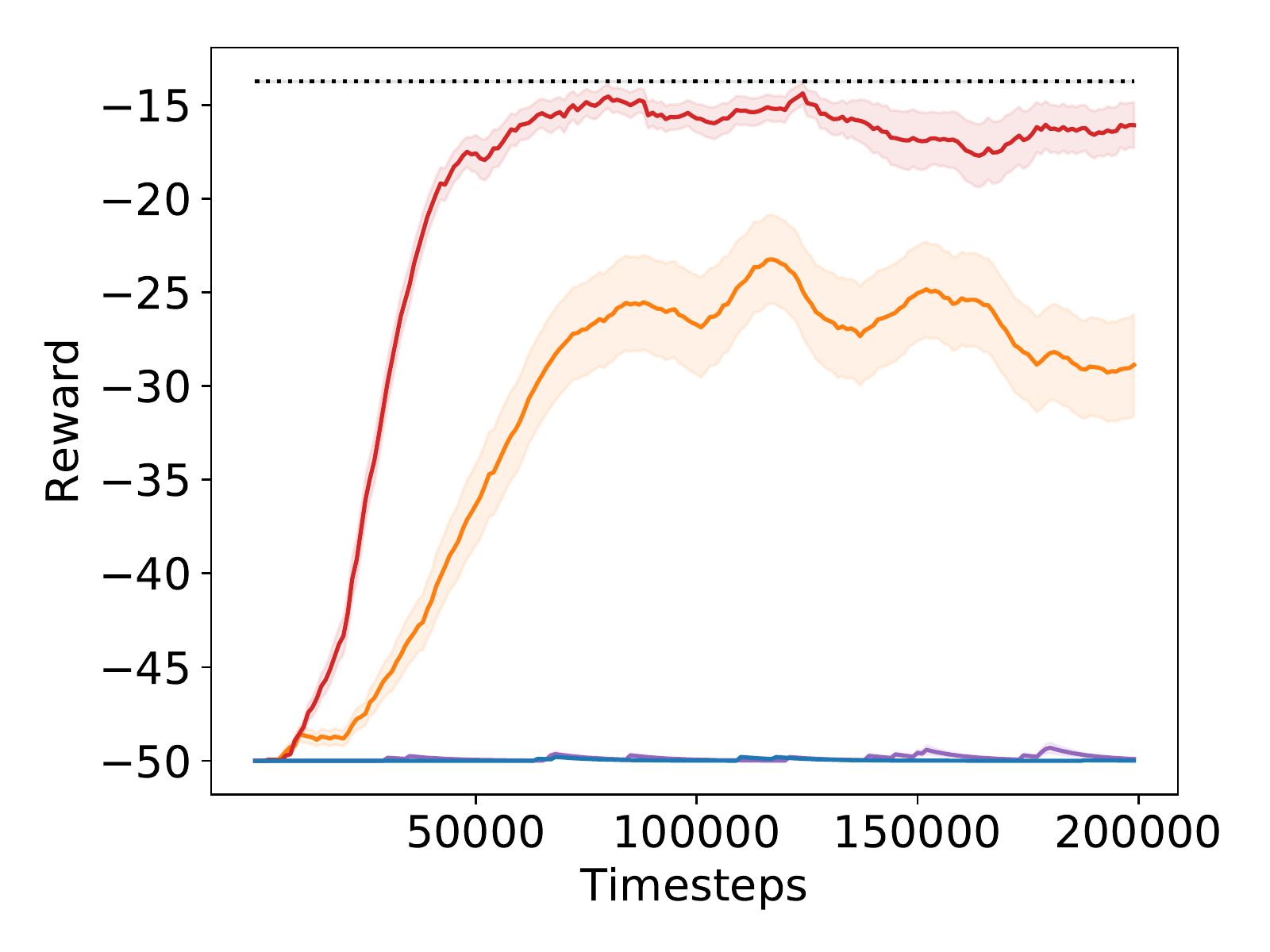}
        \label{fig:ext-result-nd}
    }
    \subfigure[Sequential Pushing]
    {
        \includegraphics[width=.3\columnwidth]{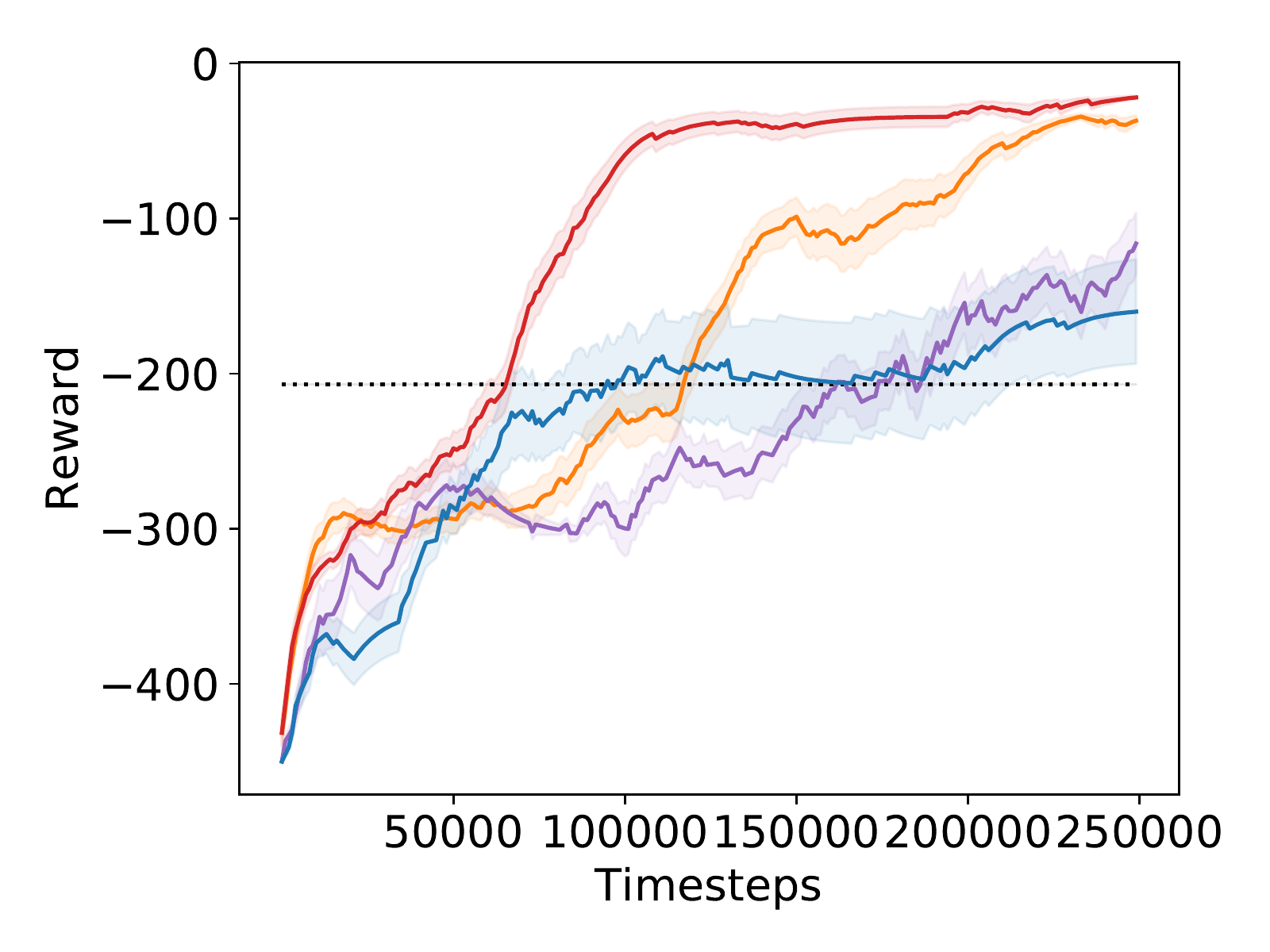}
        \label{fig:push-results-nd}
    }
    \subfigure[Door Opening]
    {
        \includegraphics[width=.3\columnwidth]{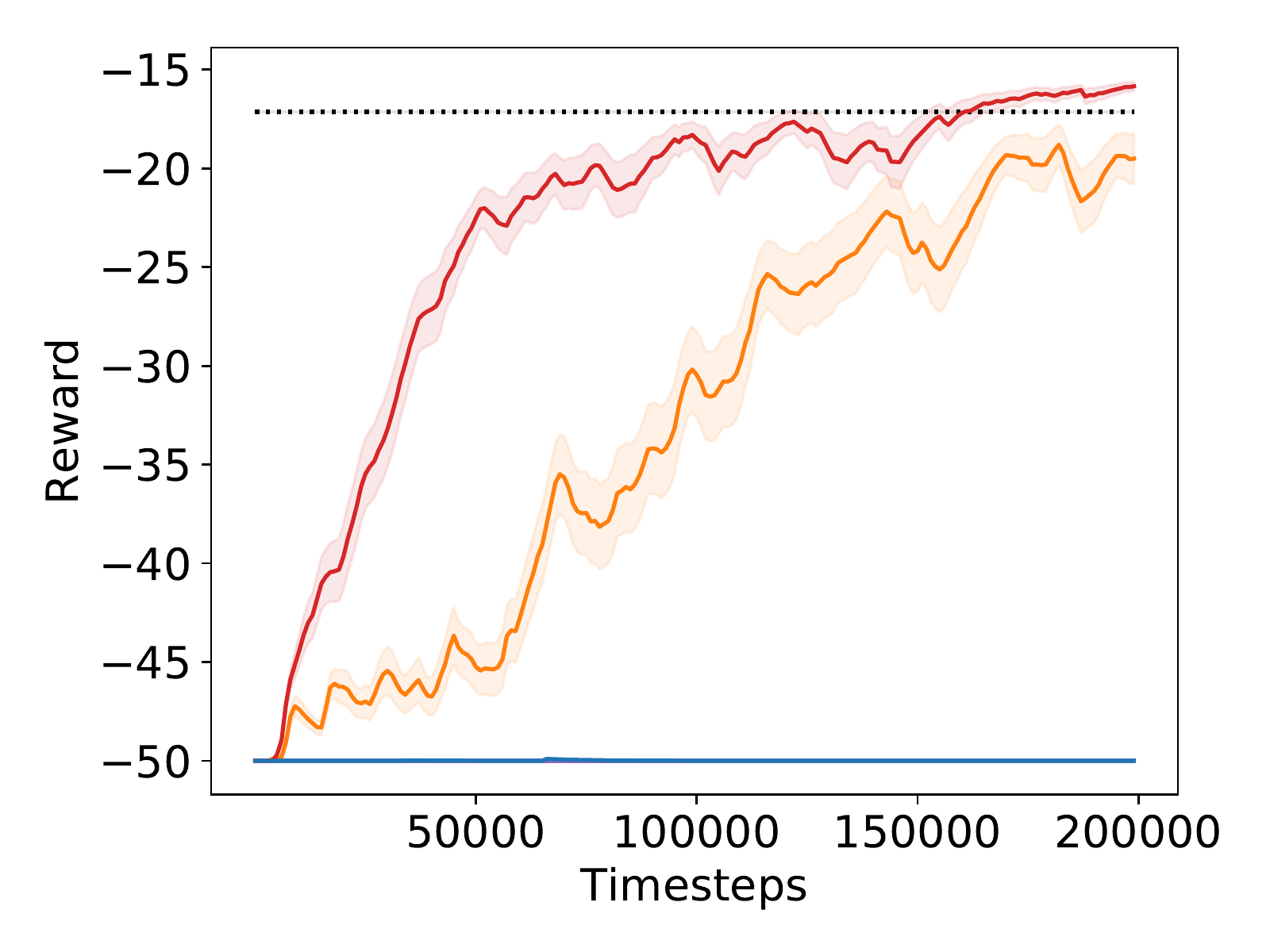}
        \label{fig:door-results-nd}
    }
    \subfigure[Block Lifting]
    {
        \includegraphics[width=.3\columnwidth]{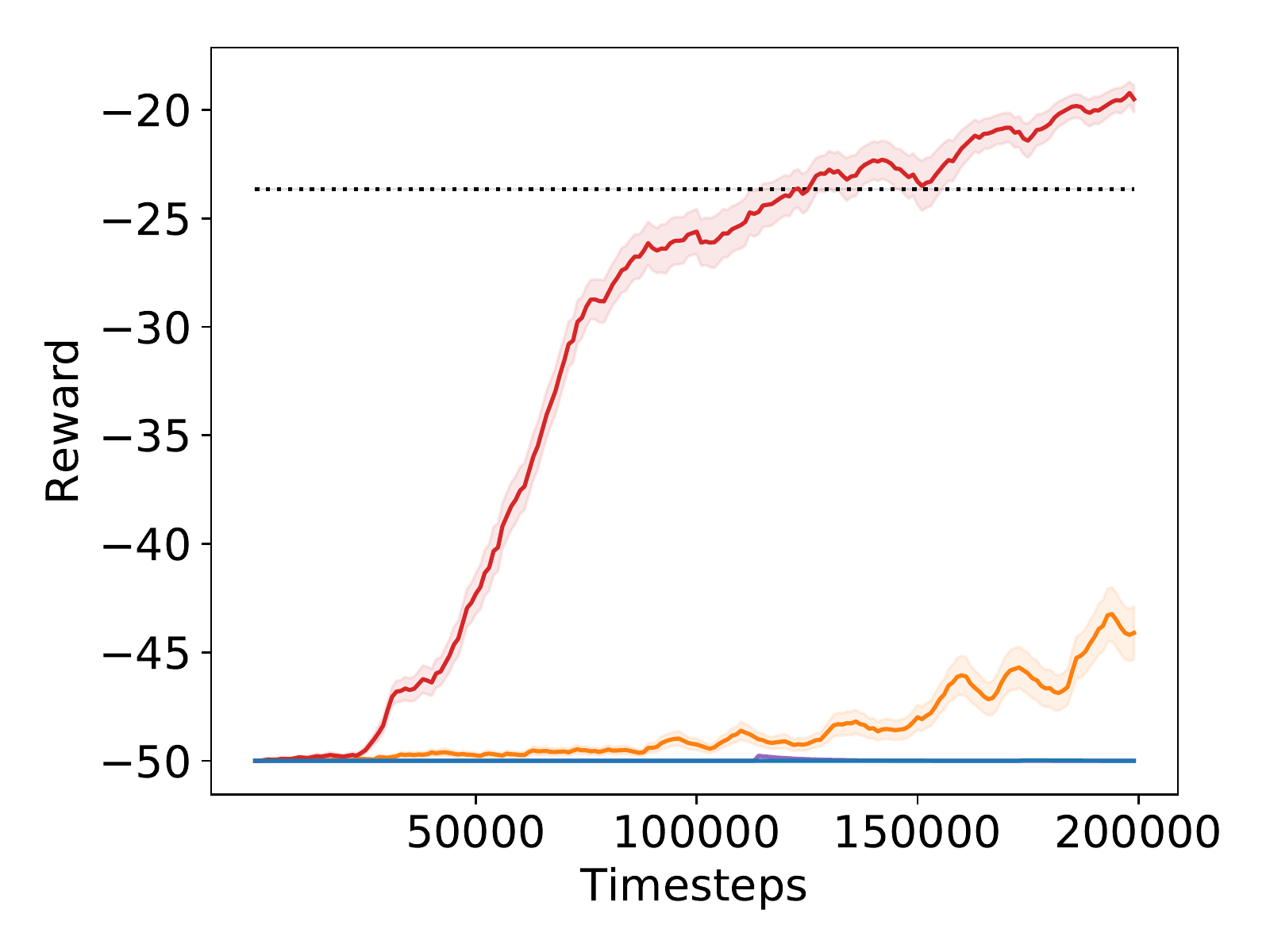}
        \label{fig:lift-results-nd}
    }
    \includegraphics[width=\columnwidth]{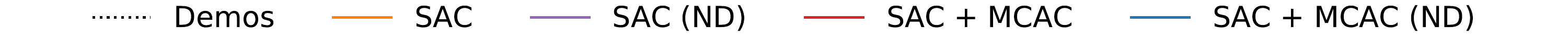}
     \caption{\new{\textbf{\algabbr{} without Demonstrations: }Learning curves showing the exponentially smoothed (smoothing factor $\gamma=0.9$) mean and standard error across 10 random seeds for experiments with demonstrations and 3 seeds for experiments without them. We see that in all environments the demonstrations are critical for learning an optimal policy. The only place the variants without demonstrations make progress is in the push environment, because of the intermediate reward for pushing each block.}}
    \label{fig:no-demos}
\end{figure*}

\new{To study the effects that \algabbr{} has without demonstration data in the replay buffer we compare performance with and without \algabbr{} and demonstrations in all environments with the SAC learner, shown in Figure~\ref{fig:no-demos}. Overall we see that, as desired, the agent is unable to make progress in most environments. The only exception is the sequential pushing environment results (Figure~\ref{fig:push-results-nd}), where the intermediate reward for pushing each block helps the agent learn to make some progress. Overall, this experiment does not conclusively answer whether \algabbr{} is helpful without demonstrations, but this is an exciting direction for future work.}

\subsection{\algabbr{} Sensitivity Experiments}
\label{subsubsec:sensitivity}
\begin{figure*}
    \centering
    \subfigure[Demonstration Quality]{
        \includegraphics[width=0.45\columnwidth]{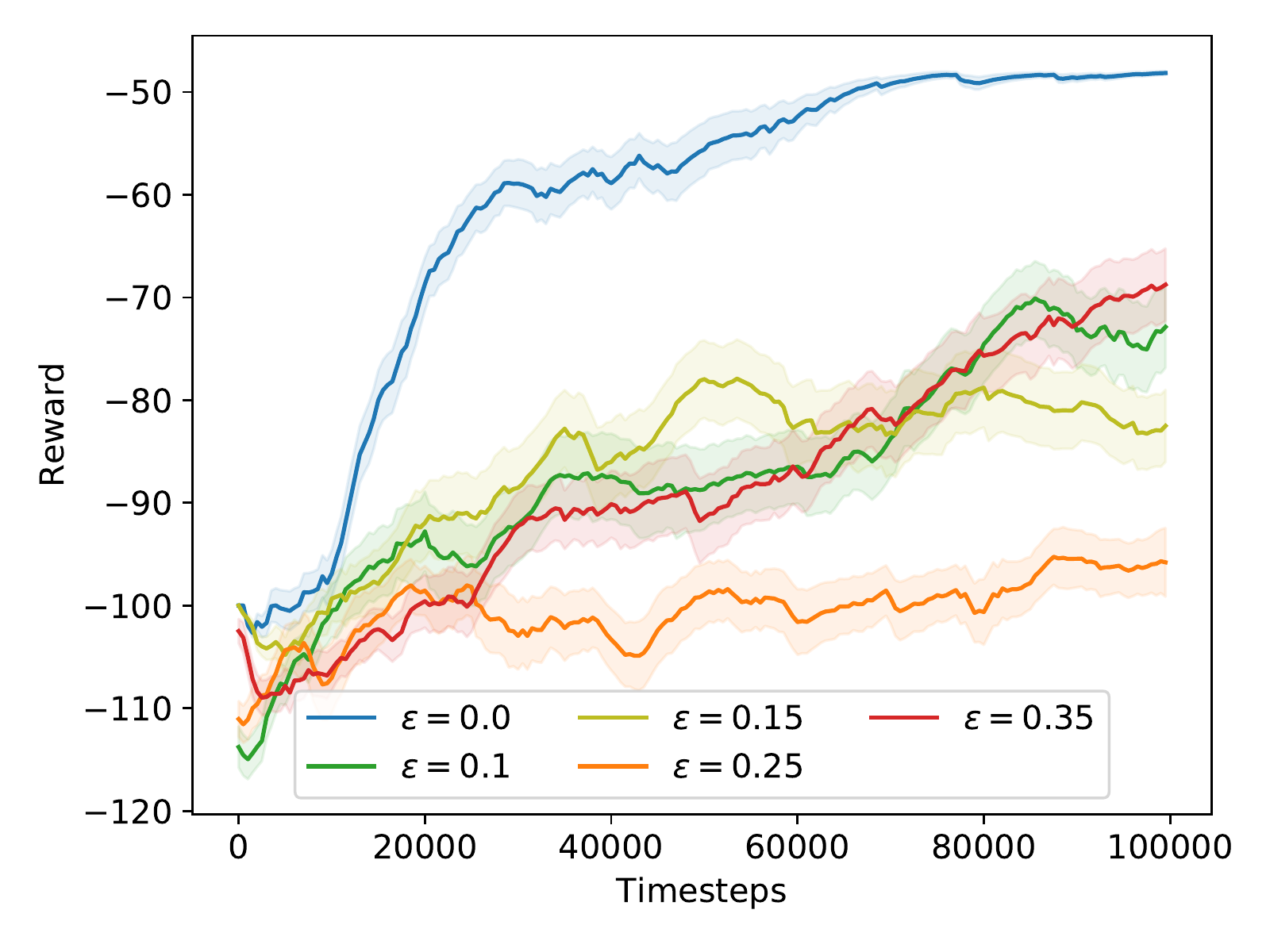}
        \label{fig:demo-quality}
    }
    \subfigure[Demonstration Quantity]{
        \includegraphics[width=0.45\columnwidth]{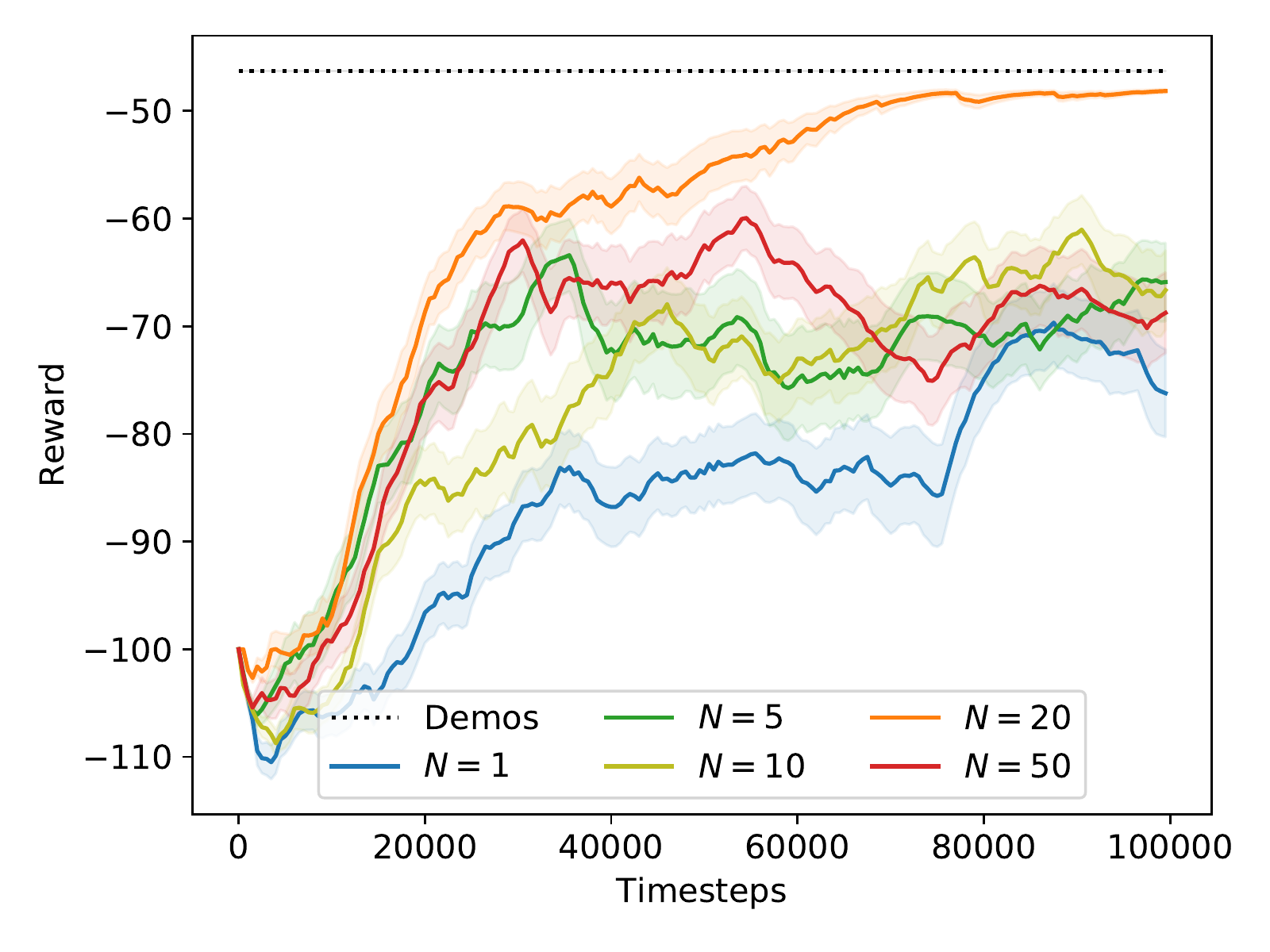}
        \label{fig:demo-quantity}
    }
    \caption{\textbf{\algabbr{} Sensitivity Experiments: }Learning curves showing the \new{exponentially smoothed (smoothing factor $\gamma=0.9$)} mean and standard error across 10 random seeds for varying demonstration qualities (a) and quantities (b) for SAC + MCAC. (a): Results suggest that \algabbr{} is somewhat sensitive to demonstration quality, as when $\epsilon$-greedy noise is injected into the demonstrator, \algabbr{}'s performance does drop significantly, although it eventually make some task progress for most values of $\epsilon$. (b): \algabbr{} appears to be much less sensitive to demonstration quantity, and is able to achieve relatively high performance even with a single task demonstration.
    }
    \label{fig:sensitivity}
\end{figure*}

In Figure~\ref{fig:sensitivity}, we first study the impact of demonstration quality (Figure~\ref{fig:demo-quality}) and quantity (Figure~\ref{fig:demo-quantity}) on \algabbr{} when applied to SAC (SAC + MCAC) on the Pointmass Navigation domain. 
We evaluate sensitivity to demonstration quality by injecting $\epsilon$-greedy noise into the demonstrator for the Pointmass Navigation domain. Results suggest that \algabbr{} is somewhat sensitive to demonstration quality, since \algabbr{}'s performance does drop significantly for most values of $\epsilon$, although it still typically makes some task progress. 
In Figure~\ref{fig:demo-quantity}, results suggest that \algabbr{} is relatively robust to the number of demonstration.


\subsection{Pretraining}

\begin{figure*}[htb!]
\centering
    \subfigure[Pointmass Navigation]
    {
        \includegraphics[width=.3\columnwidth]{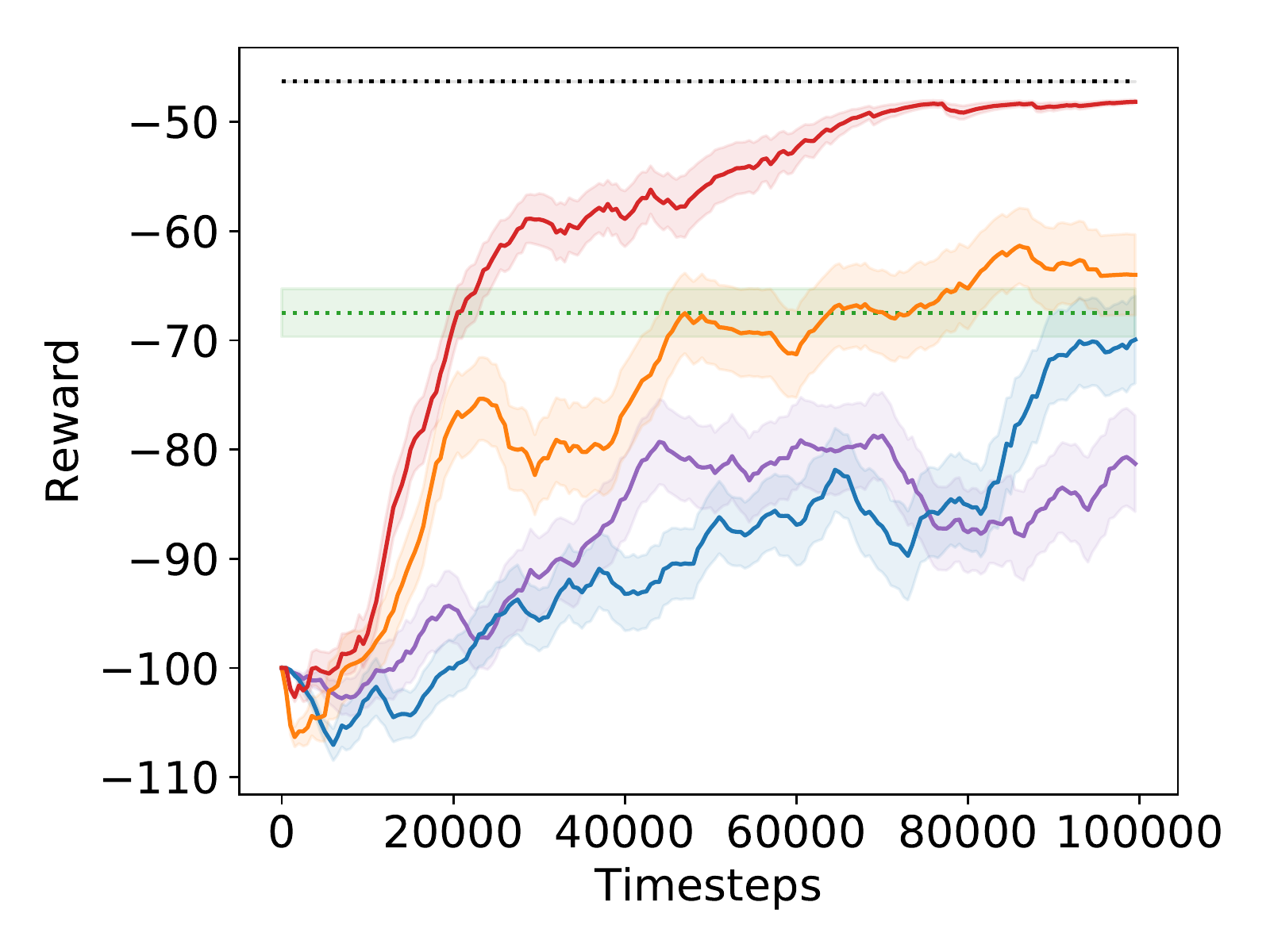}
        \label{fig:nav-results}
    }
    \subfigure[Block Extraction]
    {
        \includegraphics[width=.3\columnwidth]{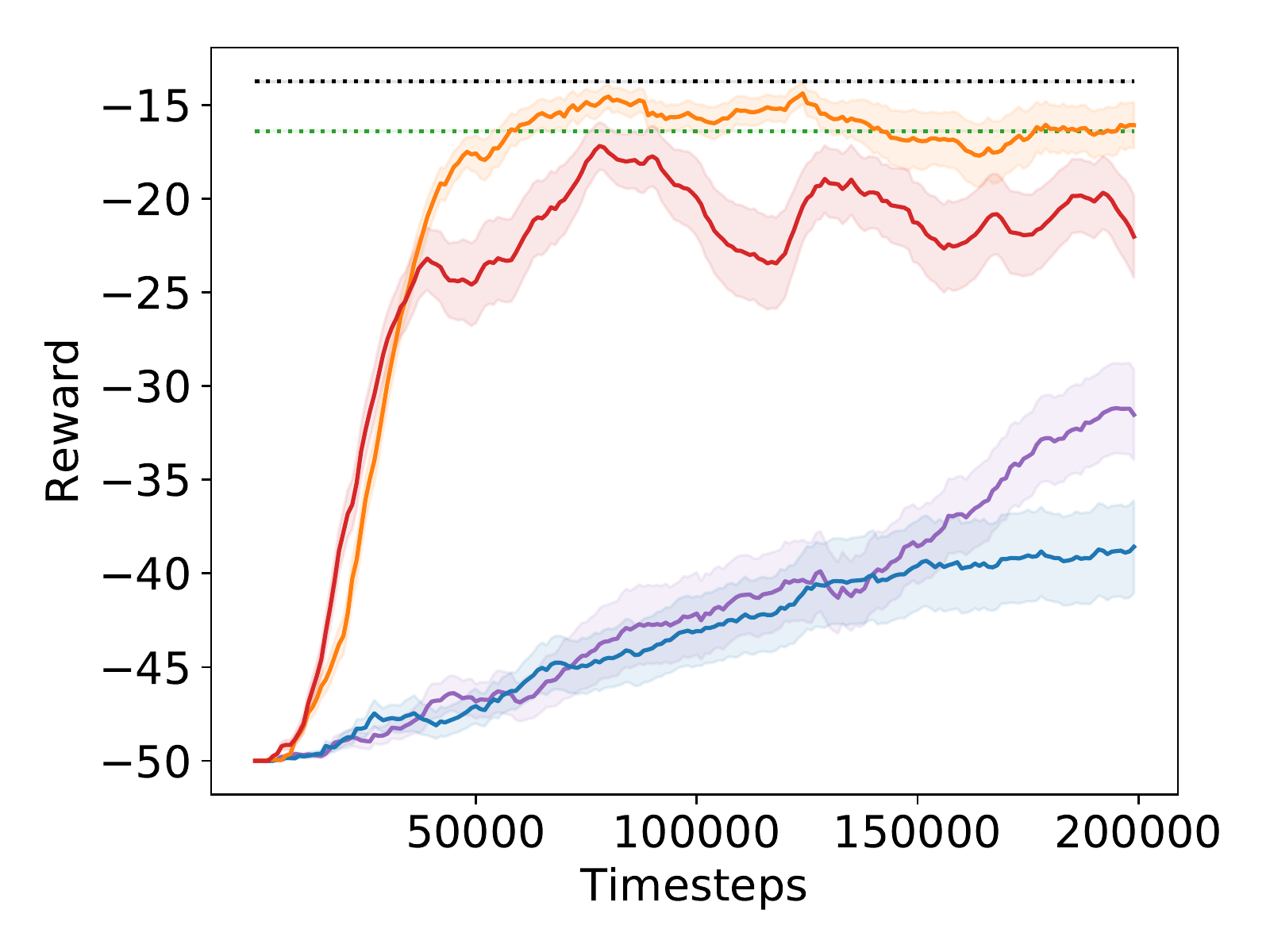}
        \label{fig:ext-result}
    }
    \subfigure[Sequential Pushing]
    {
        \includegraphics[width=.3\columnwidth]{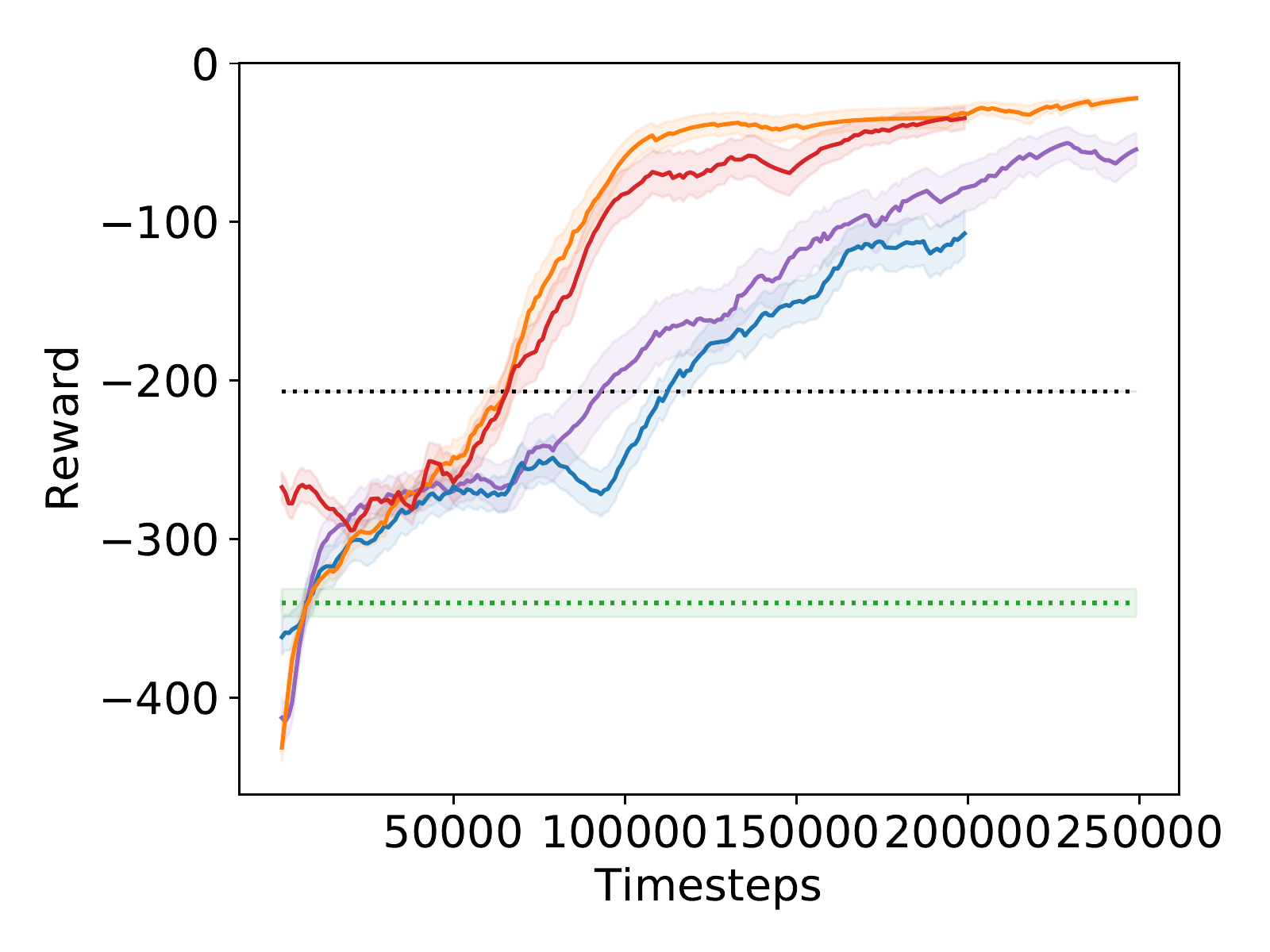}
        \label{fig:push-results}
    }
    \subfigure[Door Opening]
    {
        \includegraphics[width=.3\columnwidth]{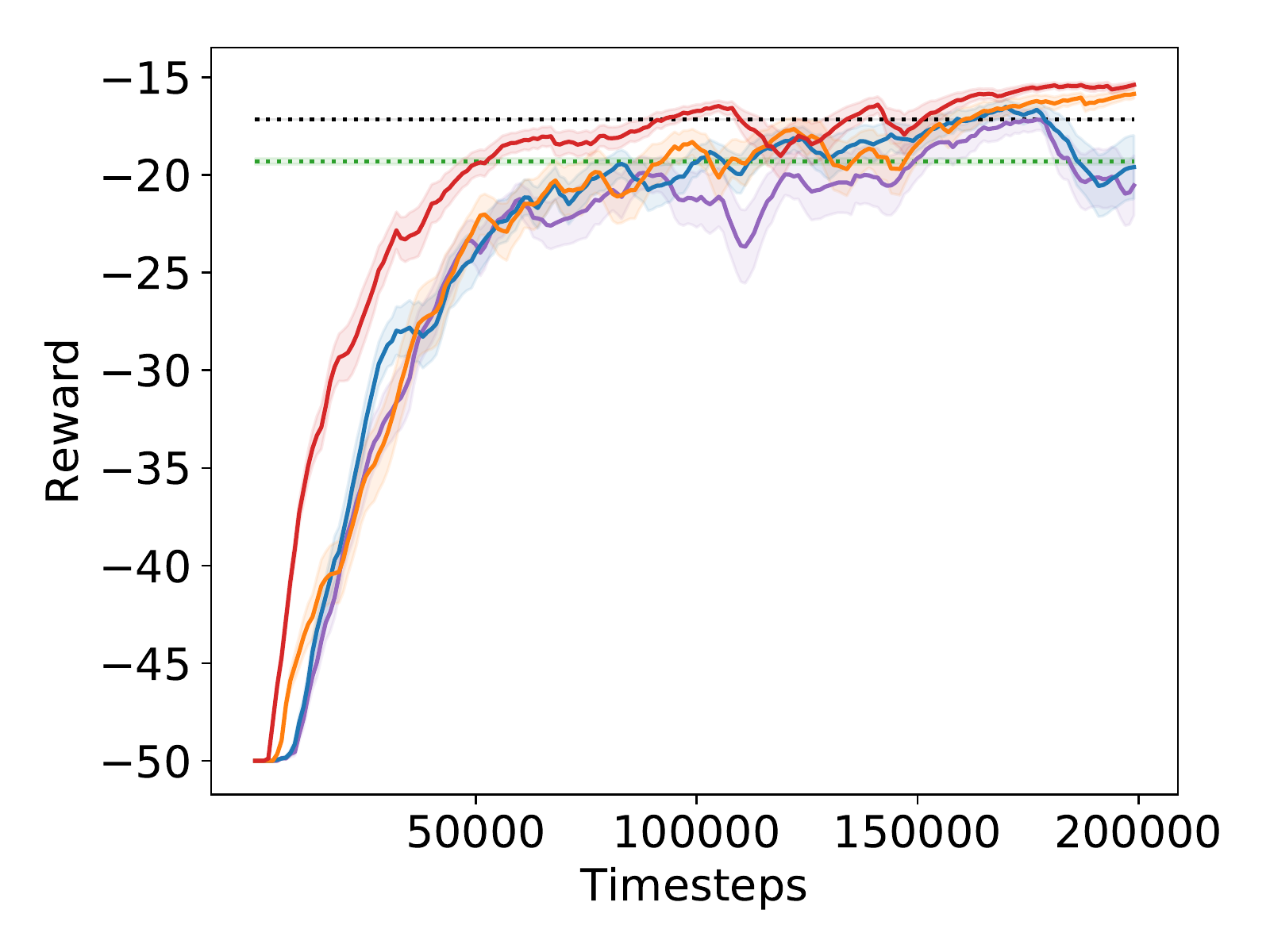}
        \label{fig:door-results}
    }
    \subfigure[Block Lifting]
    {
        \includegraphics[width=.3\columnwidth]{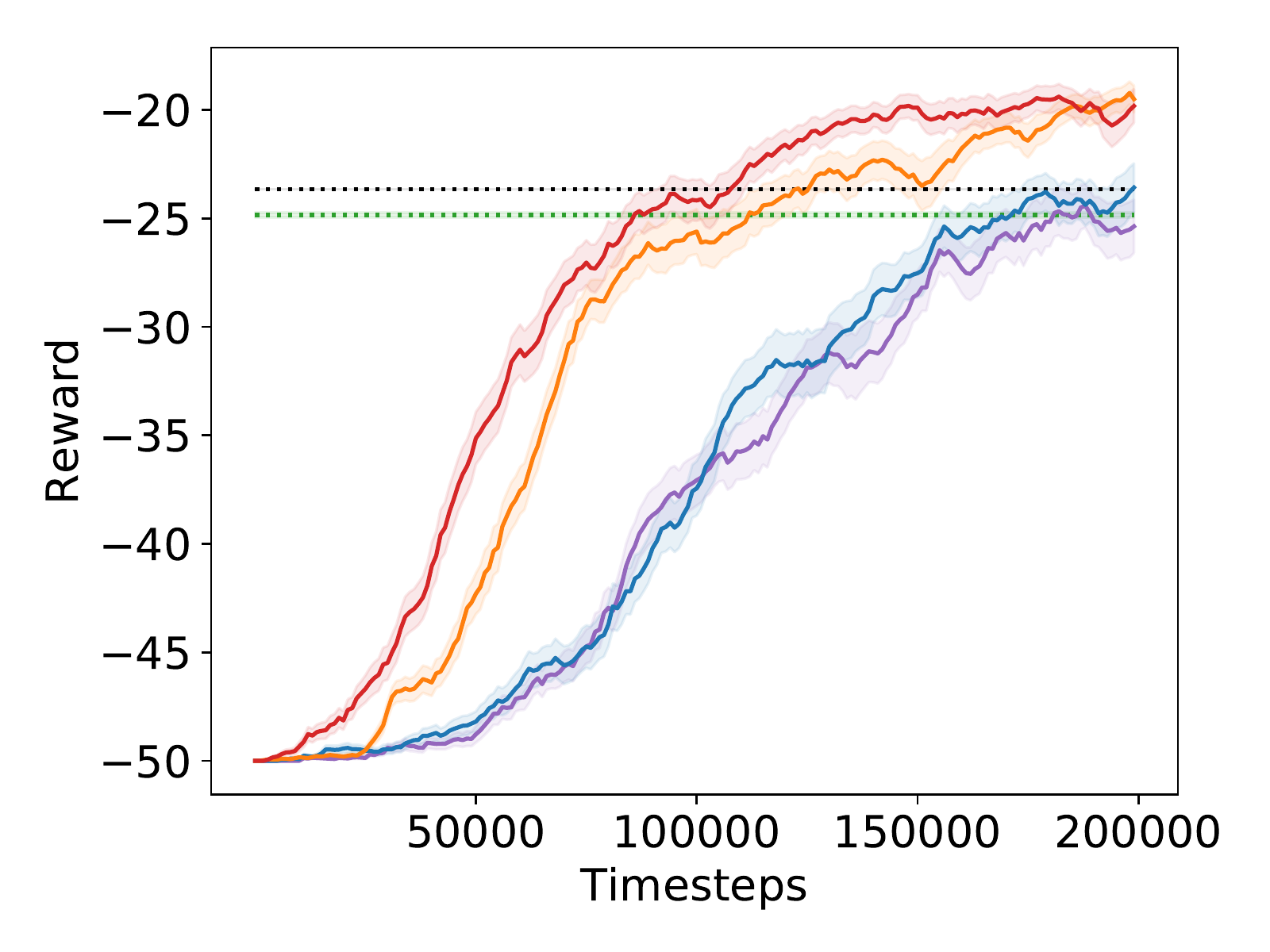}
        \label{fig:lift-results}
    }
    \includegraphics[width=\columnwidth]{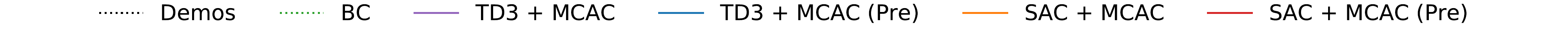}
     \caption{\textbf{\algabbr{} with and without Pretraining Results: }Learning curves showing the \new{exponentially smoothed (smoothing factor $\gamma=0.9$)} mean and standard error across 10 random seeds. We find that other than in the navigation environment pretraining does not provide a significant benefit.}
    \label{fig:pretraining-ablation}
\end{figure*}

In Figure~\ref{fig:pretraining-ablation} we ablate on the pretraining step of the algorithm to determine whether it is useful to include. We find that it was helpful for the Navigation environment but unhelpful and sometimes limiting in other settings. Thus, we leave pretraining as a hyperparameter to be tuned.

\subsection{Lambda Weighting}

\begin{figure*}[htb!]
\centering
    \subfigure[Pointmass Navigation]
    {
        \includegraphics[width=.3\columnwidth]{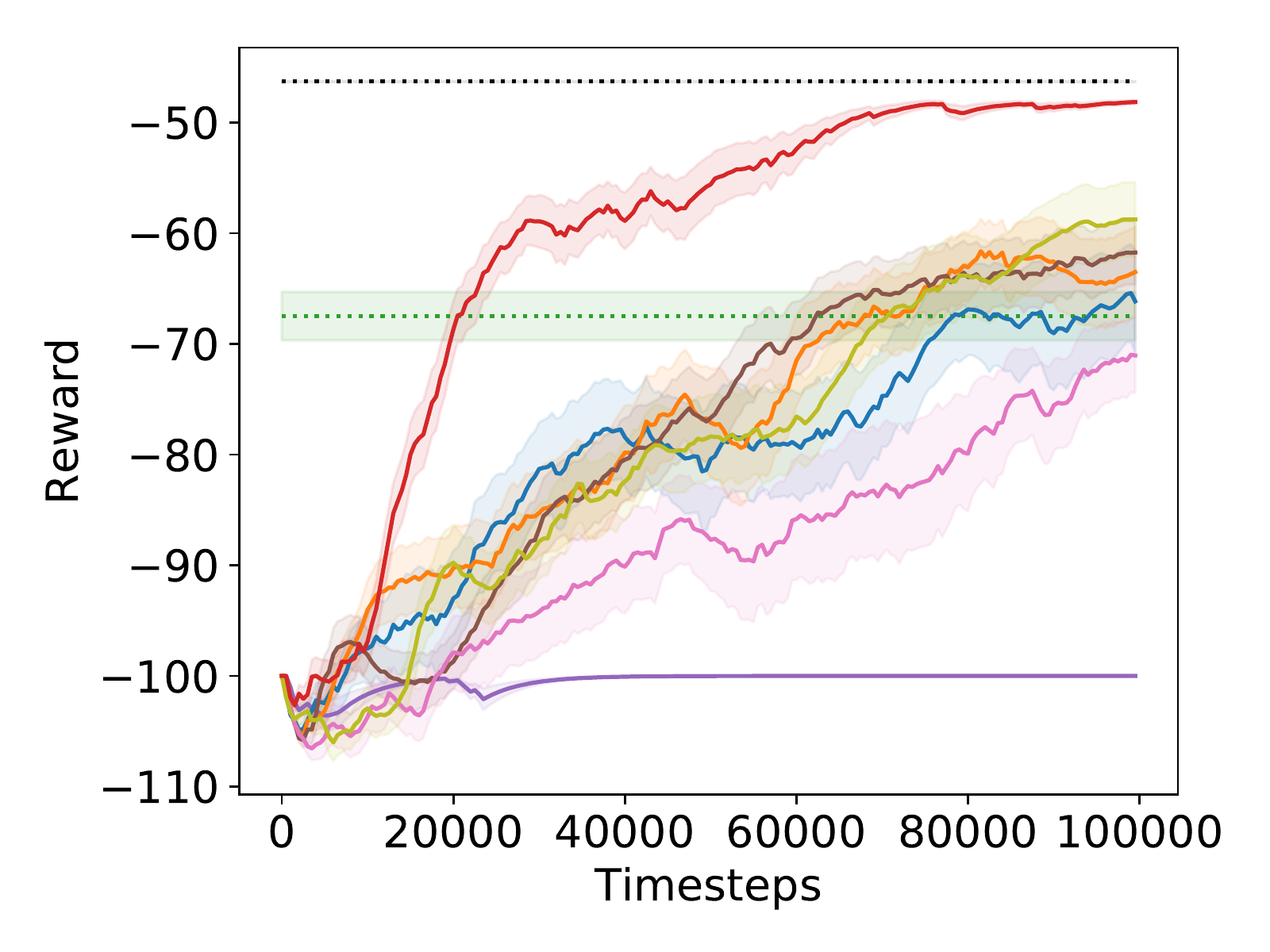}
        \label{fig:nav-results}
    }
    \subfigure[Block Extraction]
    {
        \includegraphics[width=.3\columnwidth]{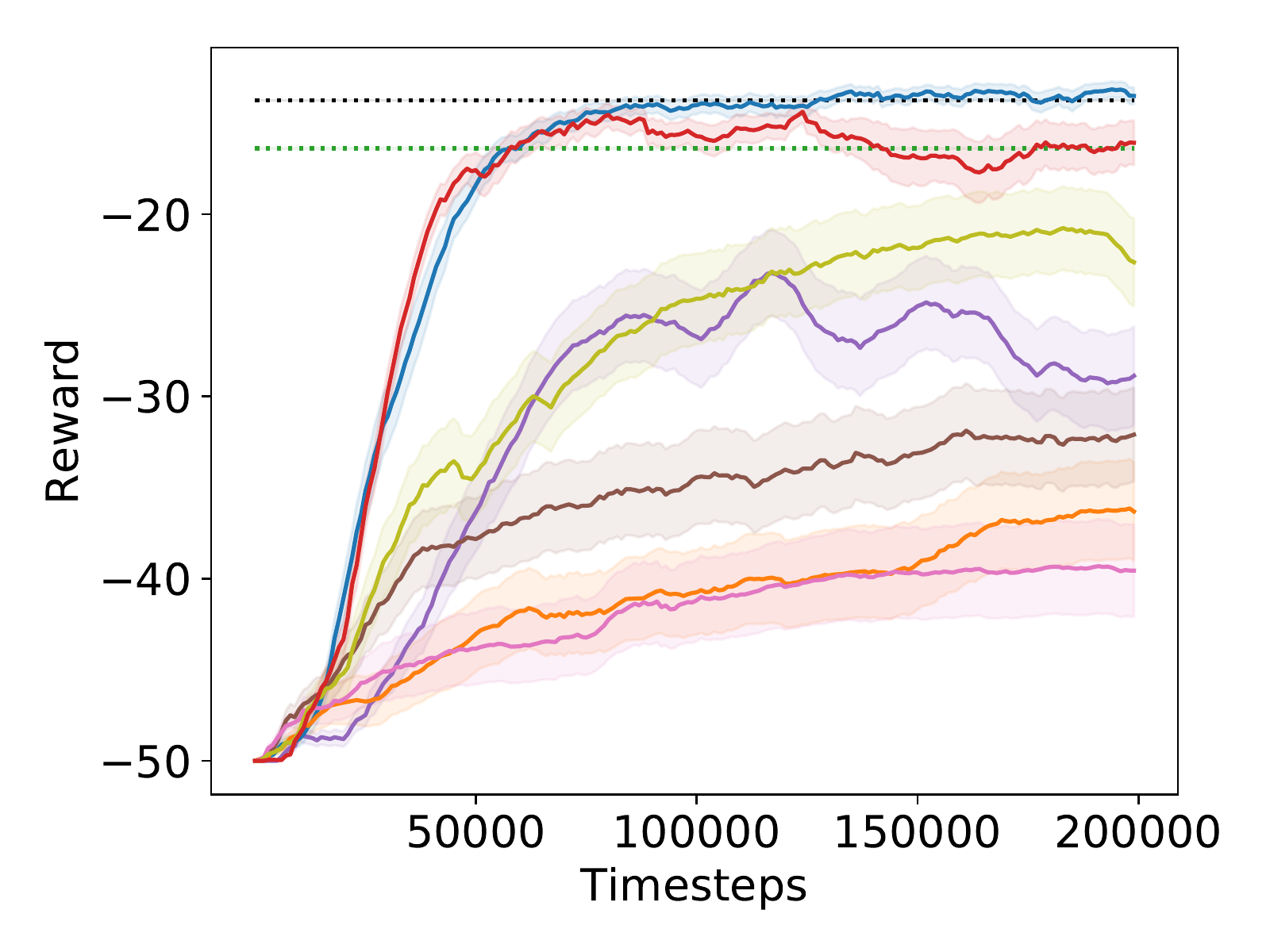}
        \label{fig:ext-result}
    }
    \subfigure[Sequential Pushing]
    {
        \includegraphics[width=.3\columnwidth]{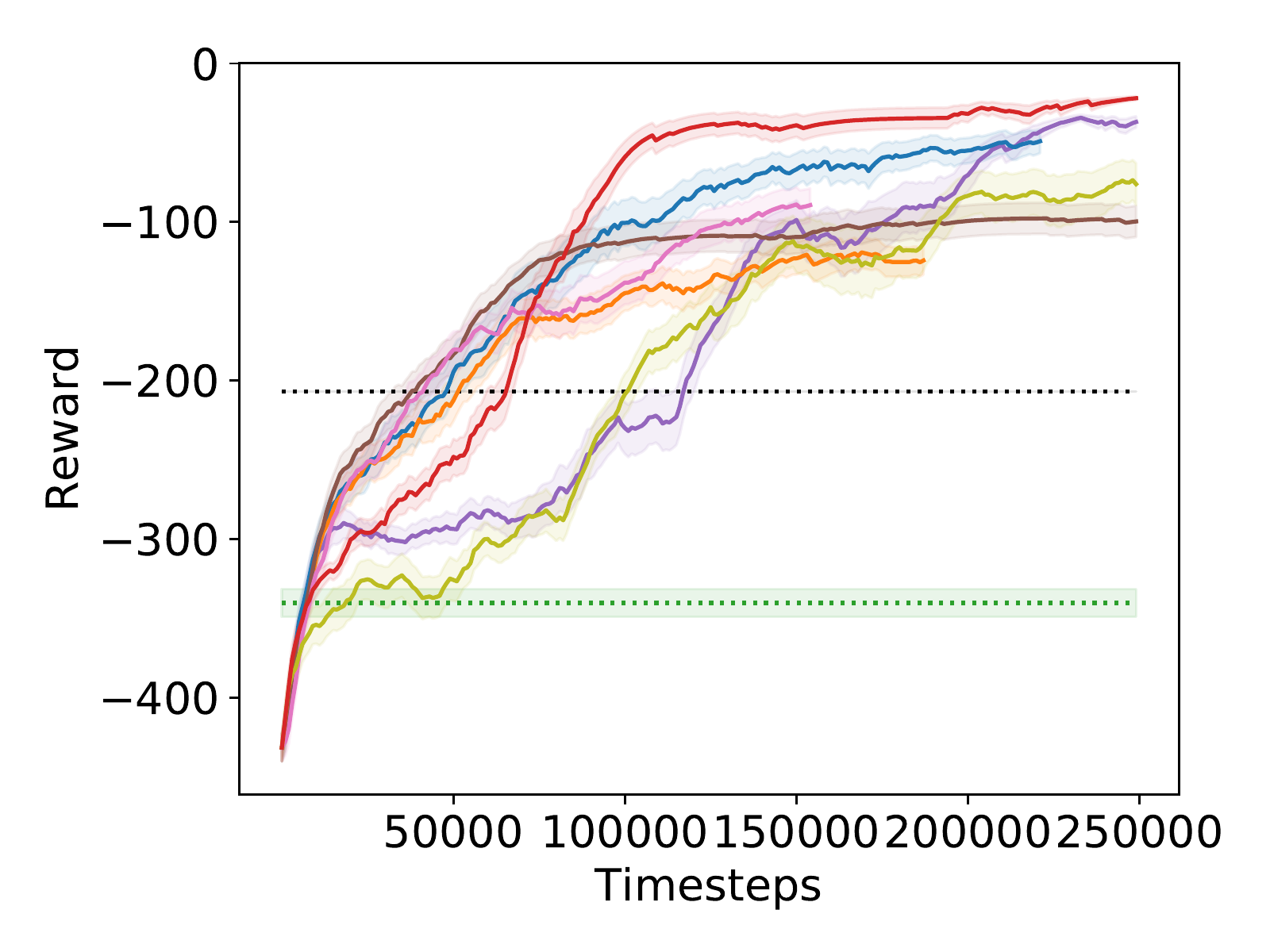}
        \label{fig:push-results}
    }
    \subfigure[Door Opening]
    {
        \includegraphics[width=.3\columnwidth]{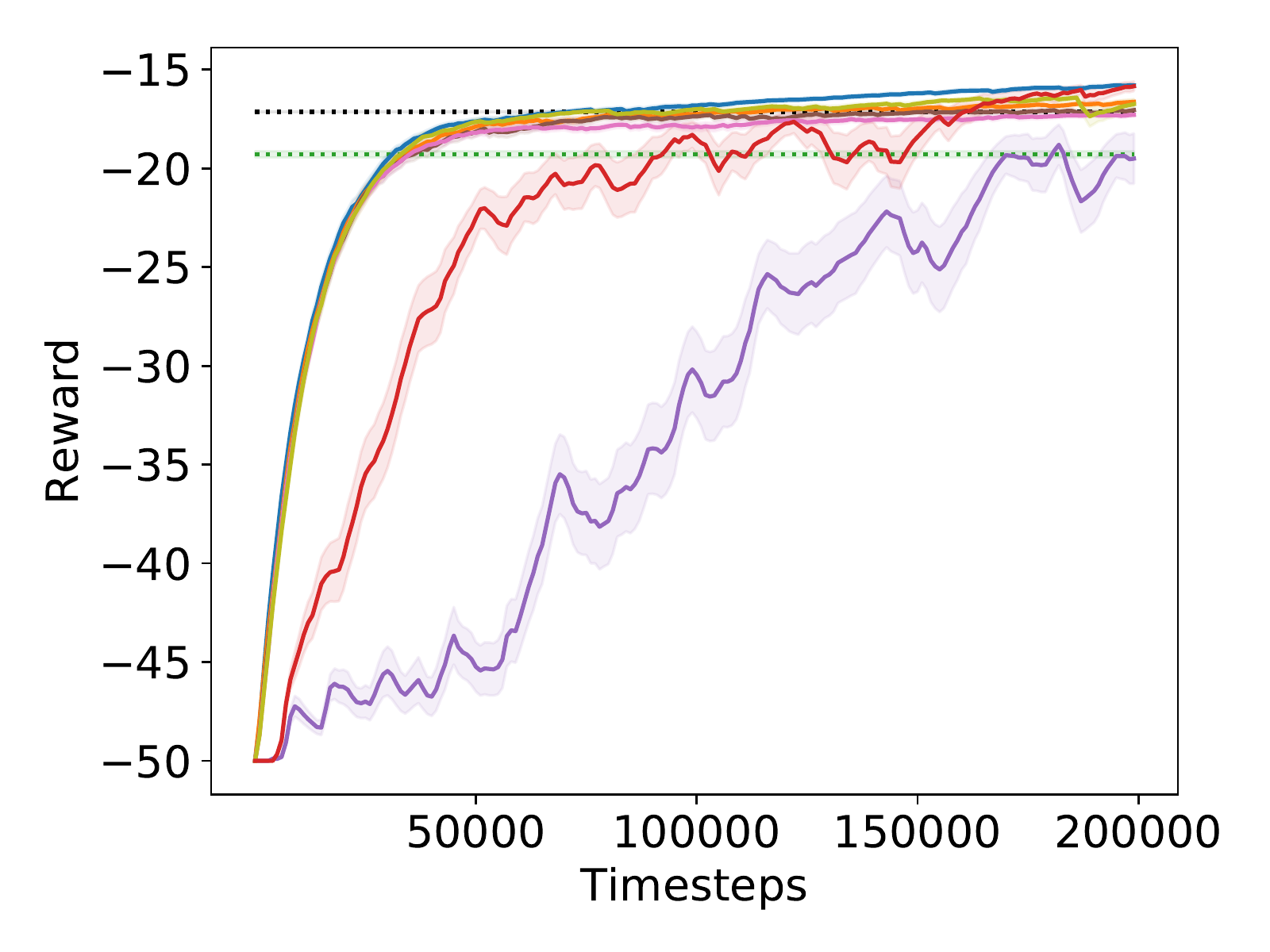}
        \label{fig:door-results}
    }
    \subfigure[Block Lifting]
    {
        \includegraphics[width=.3\columnwidth]{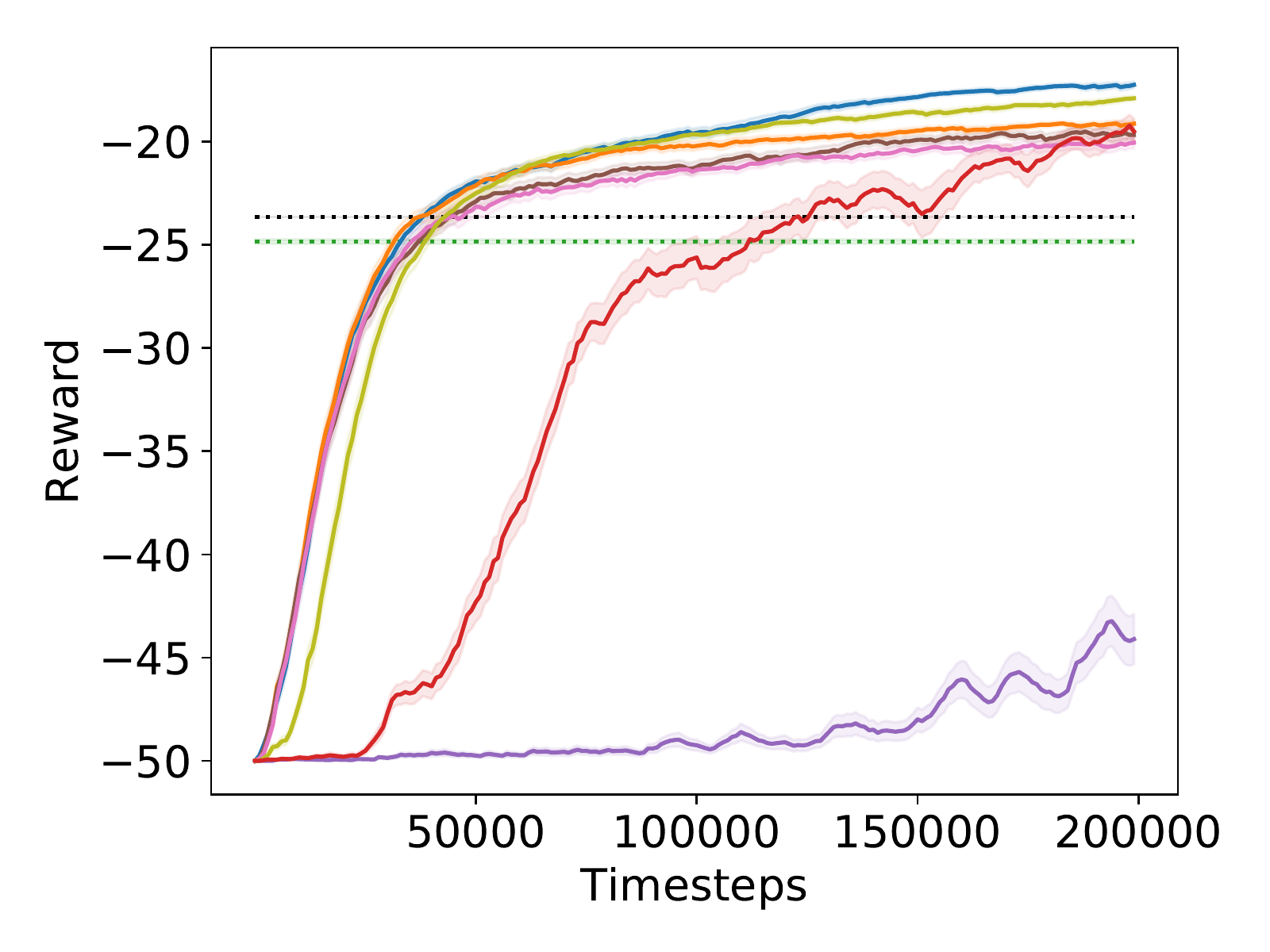}
        \label{fig:lift-results}
    }
    \includegraphics[width=\columnwidth]{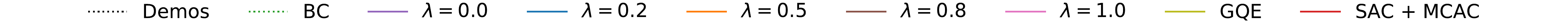}
     \caption{\textbf{\algabbr{} with Various $\lambda$ Weighting Results: }Learning curves showing the \new{exponentially smoothed (smoothing factor $\gamma=0.9$)} mean and standard error across 10 random seeds. Results suggest that this method performs similarly to GQE in most environments.}
    \label{fig:mixed-returns}
\end{figure*}

As an additional comparison we consider computing a weighted combination of the Monte Carlo and Bellman updates, using a $Q$ target $Q_\lambda^\textrm{target}$ given by
\begin{equation}
    Q^\textrm{target}_\lambda = (1-\lambda)Q^\textrm{target}(s_j^i, a_j^i) + \lambda Q^\textrm{target}_\textrm{MC-$\infty$}(s_j^i, a_j^i).
\end{equation}
Results of this experiment are shown in Figure~\ref{fig:mixed-returns} alongside results from our GQE experiments. We find that this method is comparable to GQE in most environments.


\subsection{Critic Tail} 
\label{sec:critic-tail}
\new{In this experiment, we consider replacing $Q^\textrm{target}_\textrm{MC-$\infty$}$ defined in the main text with a new target, which replaces the infinite geometric sum with the critic's $Q$ value estimate. Formally, for a trajectory given by $\left[(s_i, a_t, r_i), (s_{i+1}, a_{i+1}, r_{i+1}), \ldots, (s_T, a_T, r_T), s_{T+1} \right]$, a delayed target network $Q^\pi_{\theta'}$ and a policy $\pi$, the new target is}

\begin{equation}
    \label{eq:mc-target-infinite-critic-tail}
    \hat{Q}^{\textrm{target}}_\textrm{MC-$\infty$}(s_i, a_i) = \gamma^{T-j+1}Q^\pi_{\theta'}(s_{T+1}, \pi(s_{T+1})) + \sum^T_{k=j} \gamma^{k-j}r_k.
\end{equation}

\new{We compare this version to the standard version of \algabbr{} presented in the paper in Figure~\ref{fig:critic-tail}. Results suggest that the version presented in the main paper performs at least as well if not better than this version. Since this version also adds implementation difficulty and computation load, we choose to use the original $Q^\textrm{target}_\textrm{MC-$\infty$}$ estimate. However, it would be interesting in future work to study whether there are environments where this estimate has better performance.}

\begin{figure*}[htb!]
\centering
    \subfigure[Pointmass Navigation]
    {
        \includegraphics[width=.3\columnwidth]{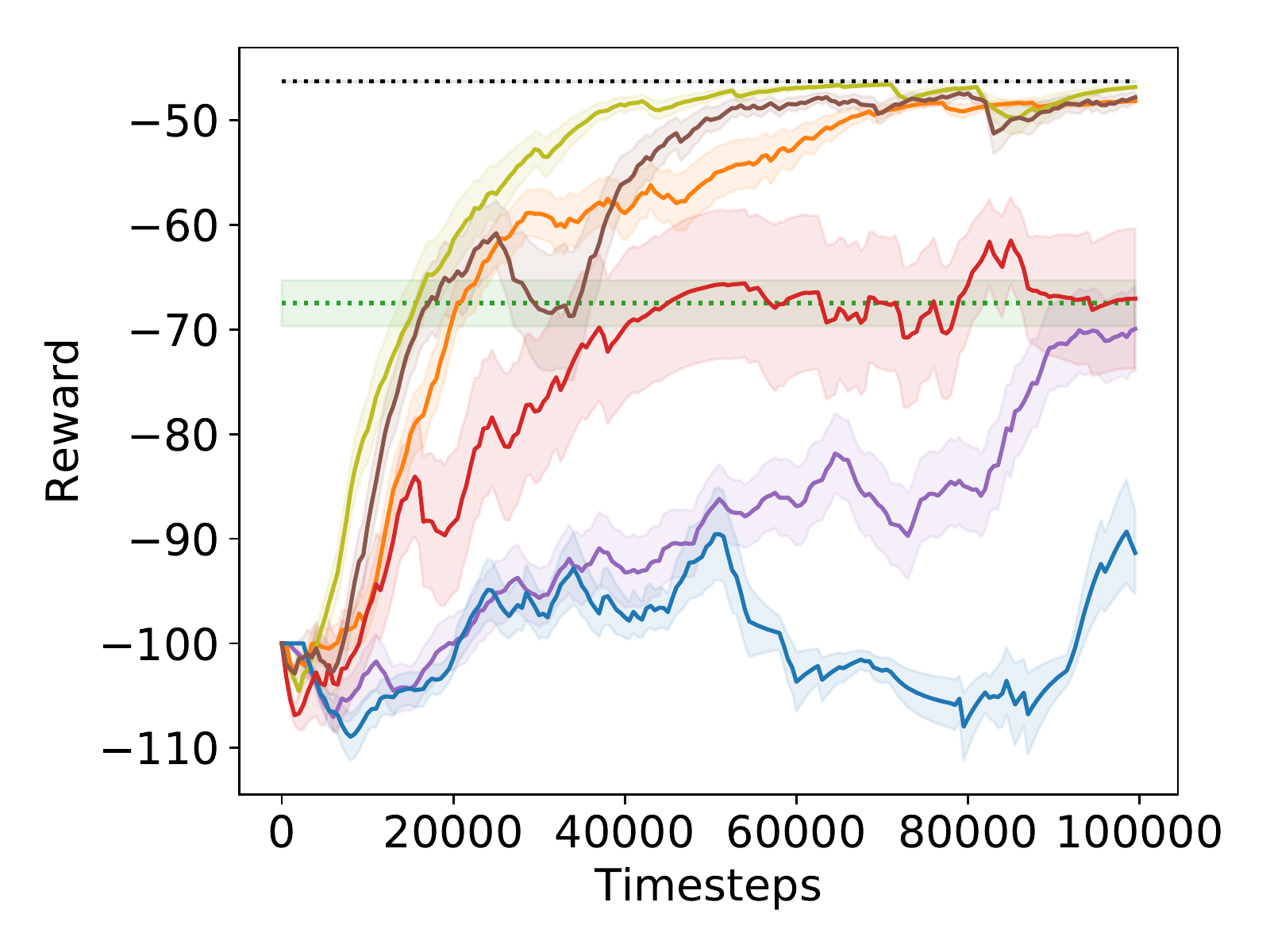}
        \label{fig:nav-results}
    }
    \subfigure[Block Extraction]
    {
        \includegraphics[width=.3\columnwidth]{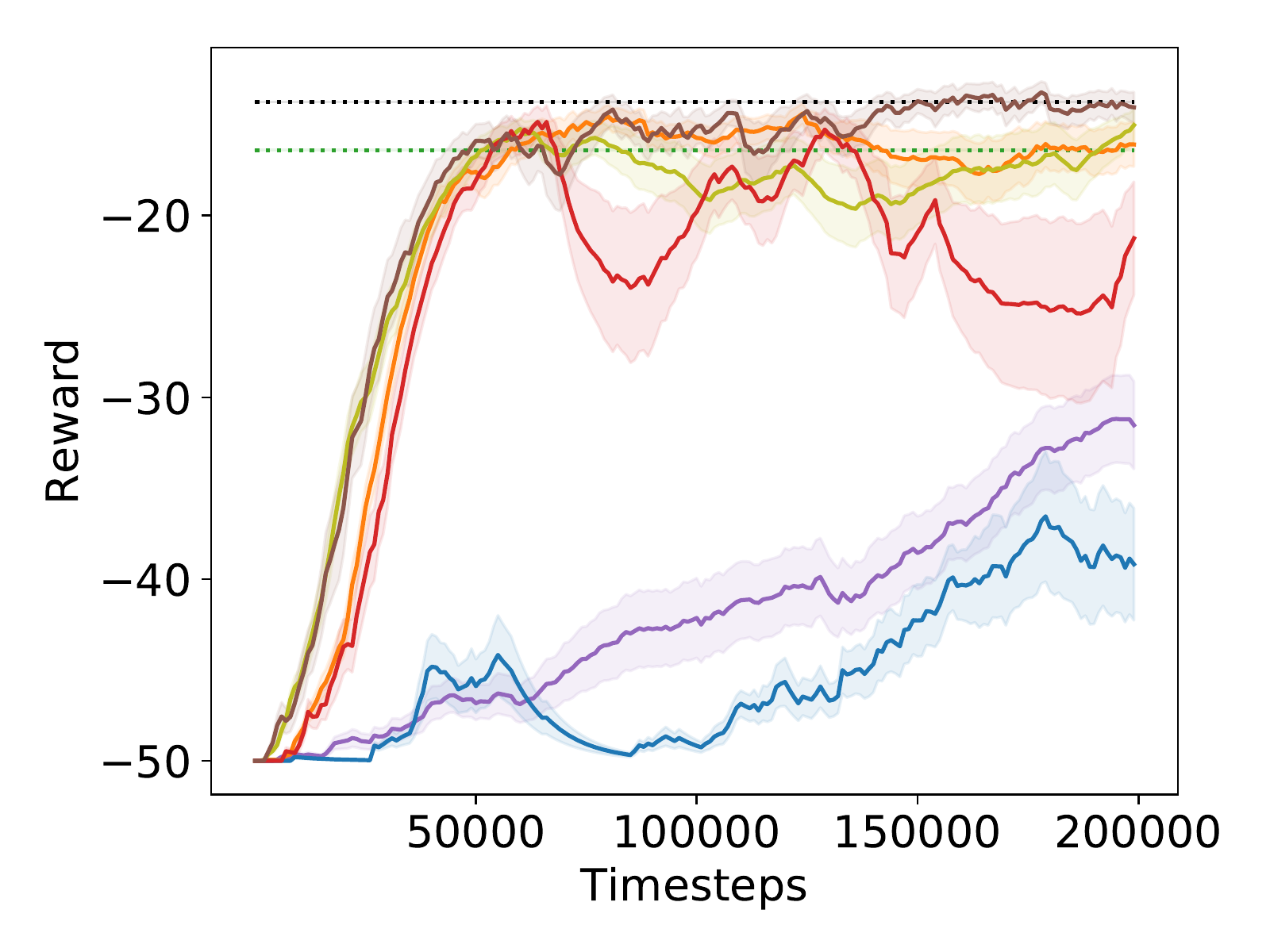}
        \label{fig:ext-result}
    }
    \subfigure[Sequential Pushing]
    {
        \includegraphics[width=.3\columnwidth]{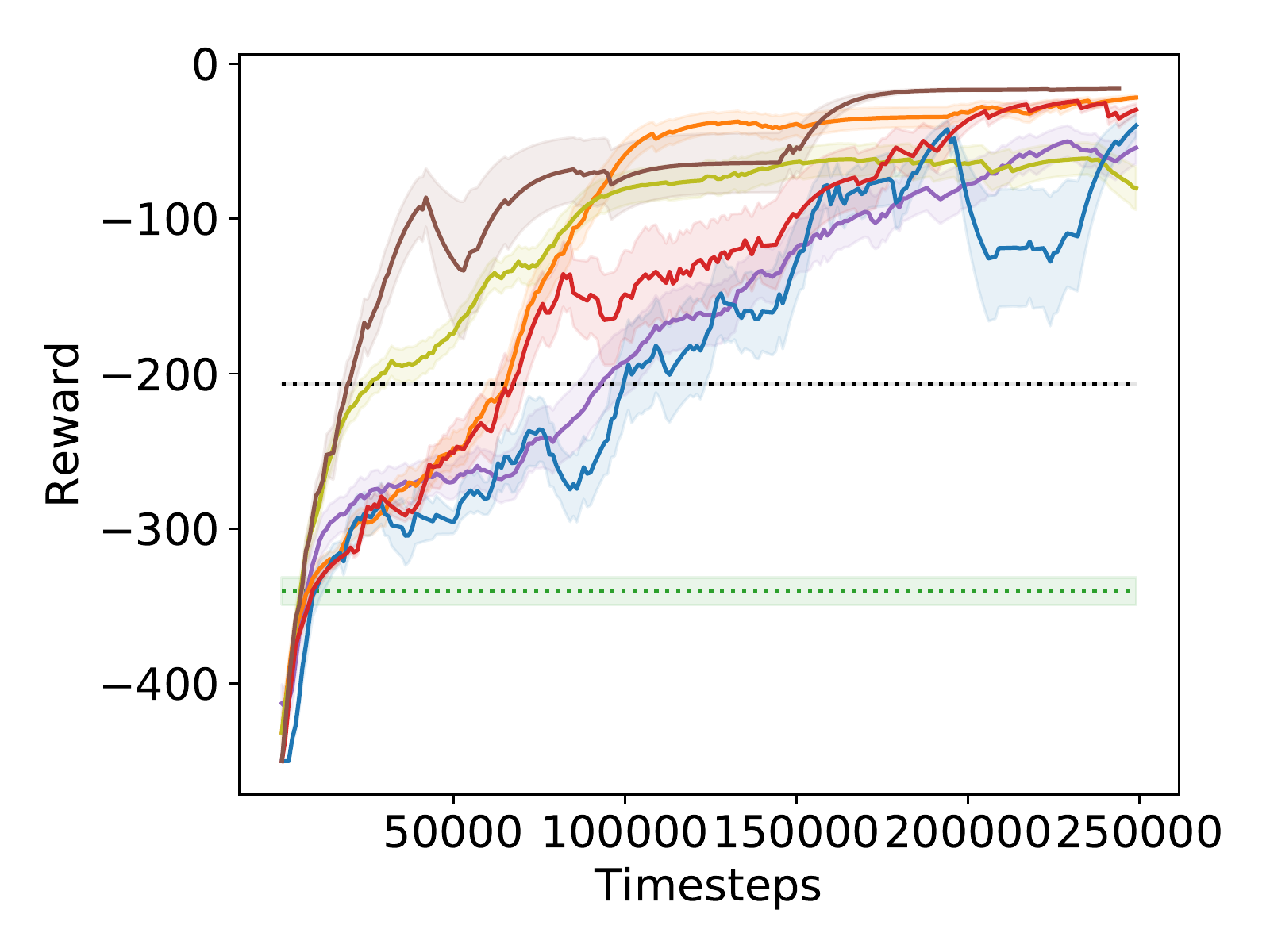}
        \label{fig:push-results}
    }
    \subfigure[Door Opening]
    {
        \includegraphics[width=.3\columnwidth]{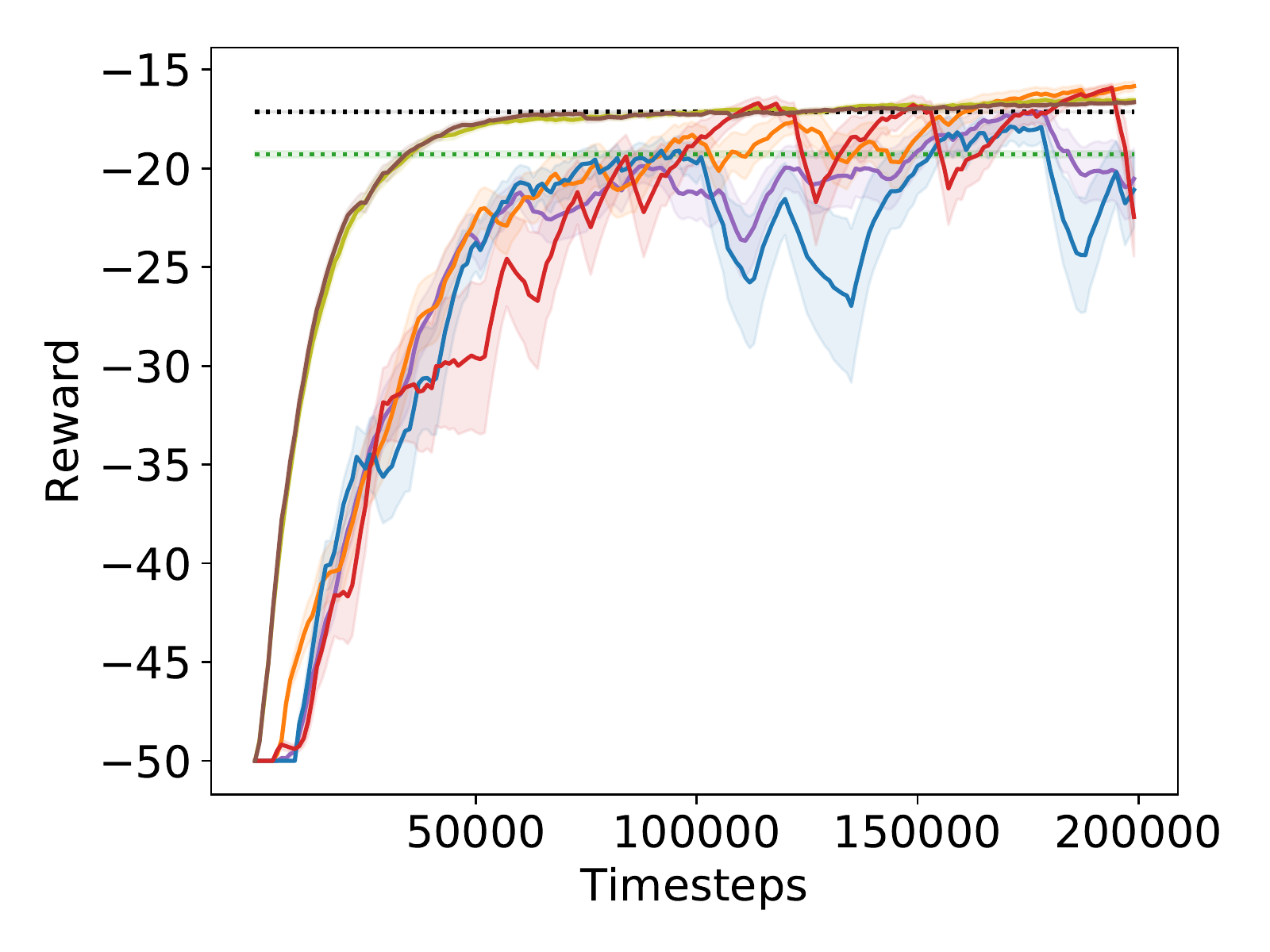}
        \label{fig:door-results}
    }
    \subfigure[Block Lifting]
    {
        \includegraphics[width=.3\columnwidth]{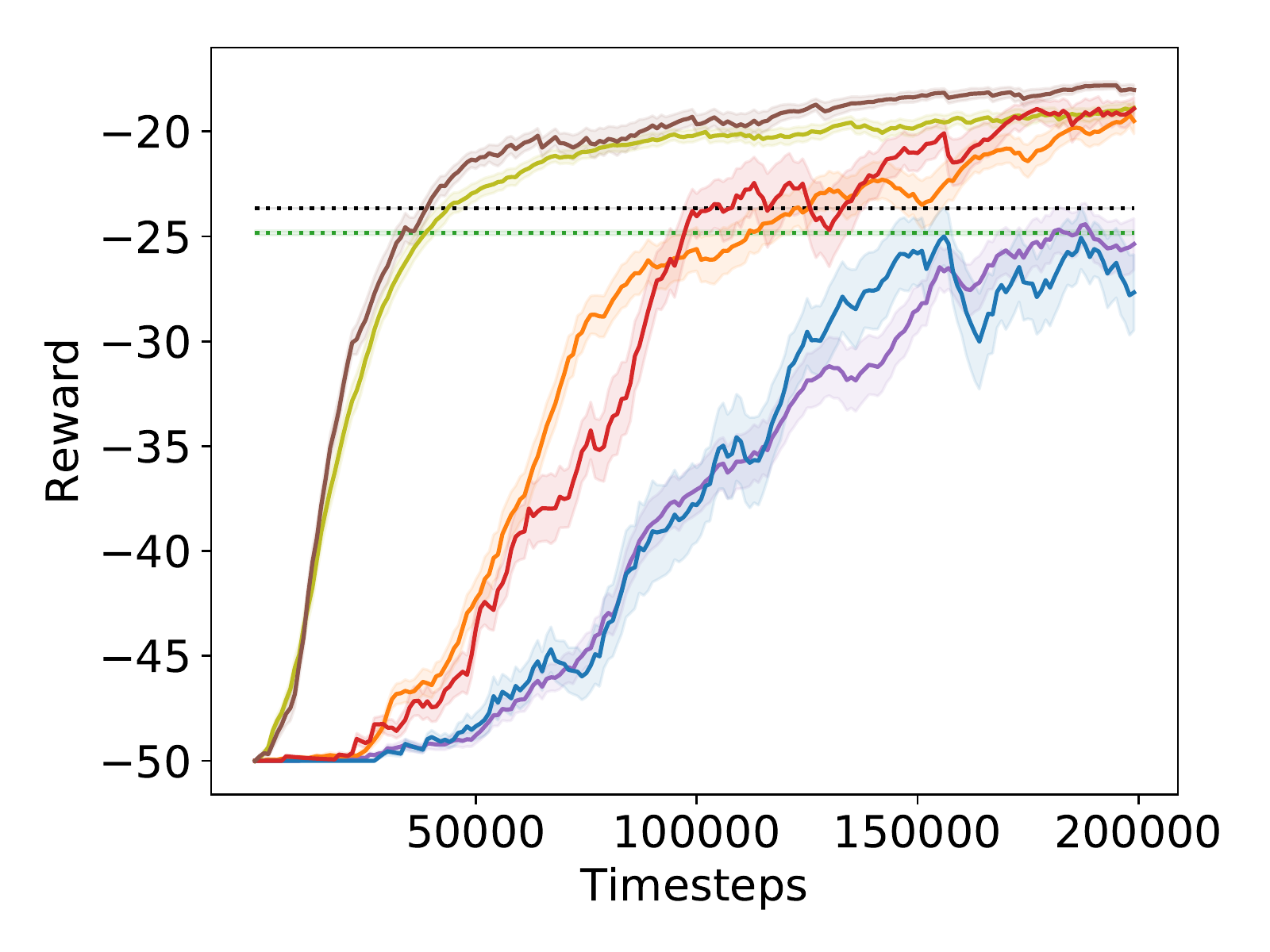}
        \label{fig:lift-results}
    }
    \includegraphics[width=\columnwidth]{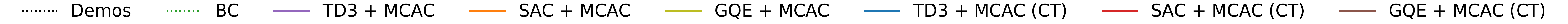}
     \caption{\new{\textbf{\algabbr{} with Critic Tail: }Learning curves showing the \new{exponentially smoothed (smoothing factor $\gamma=0.9$)} mean and standard error across 10 random seeds for experiments with the original $Q^\textrm{target}_\textrm{MC-$\infty$}$ estimate and 3 seeds for experiments with the new version described in Equation~\ref{eq:mc-target-infinite-critic-tail}. We see that in all environments the original version is at least as good as the the new version, and occasionally it performs better.}}
    \label{fig:critic-tail}
\end{figure*}

\subsection{$Q$ Estimate Scale Analysis}
\begin{figure*}[htb!]
\centering
    \subfigure[Successful Trajectories]
    {
        \includegraphics[width=.4\columnwidth]{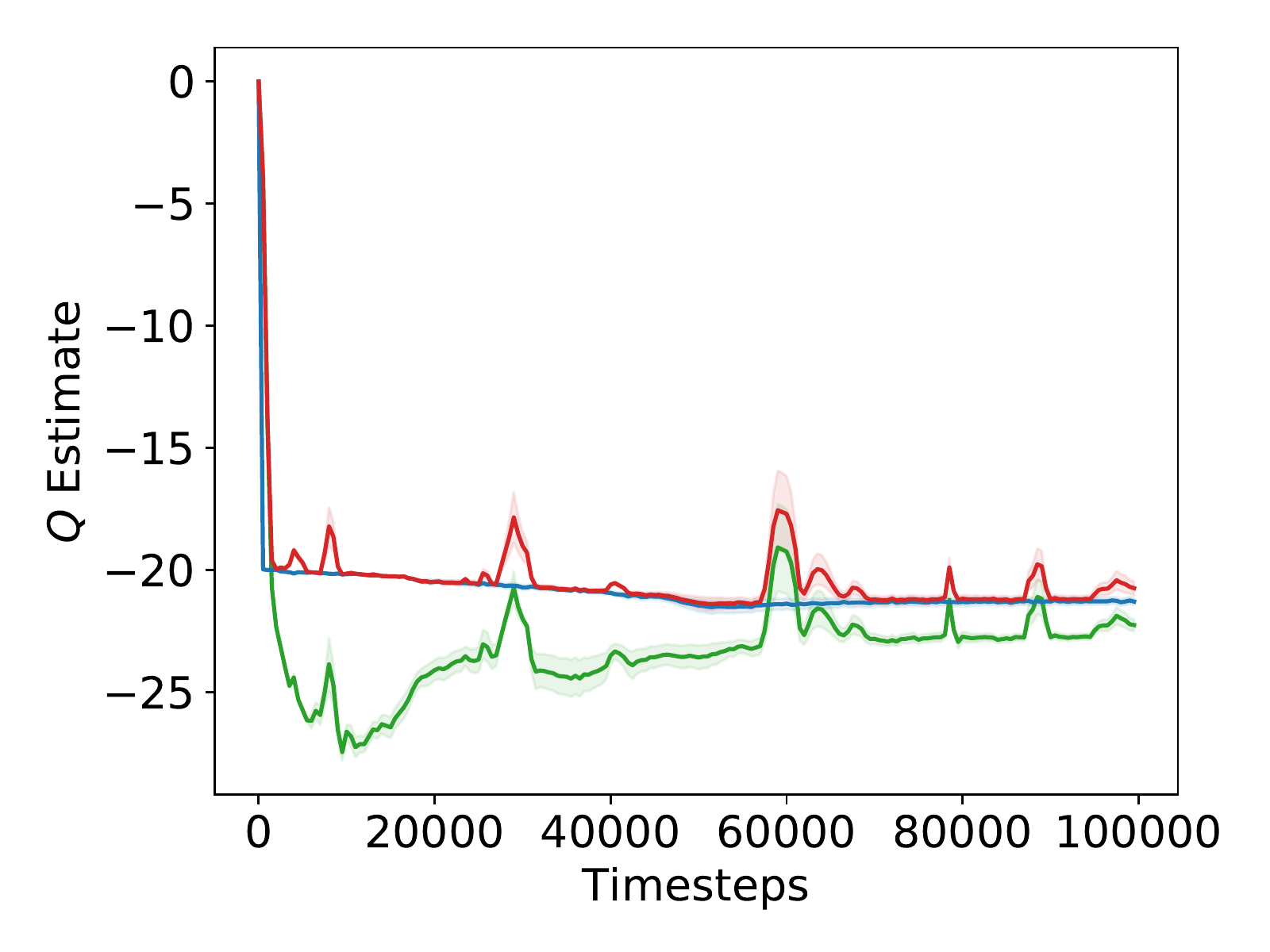}
        \label{fig:succ-scale}
    }
    \subfigure[Failed Trajectories]
    {
        \includegraphics[width=.4\columnwidth]{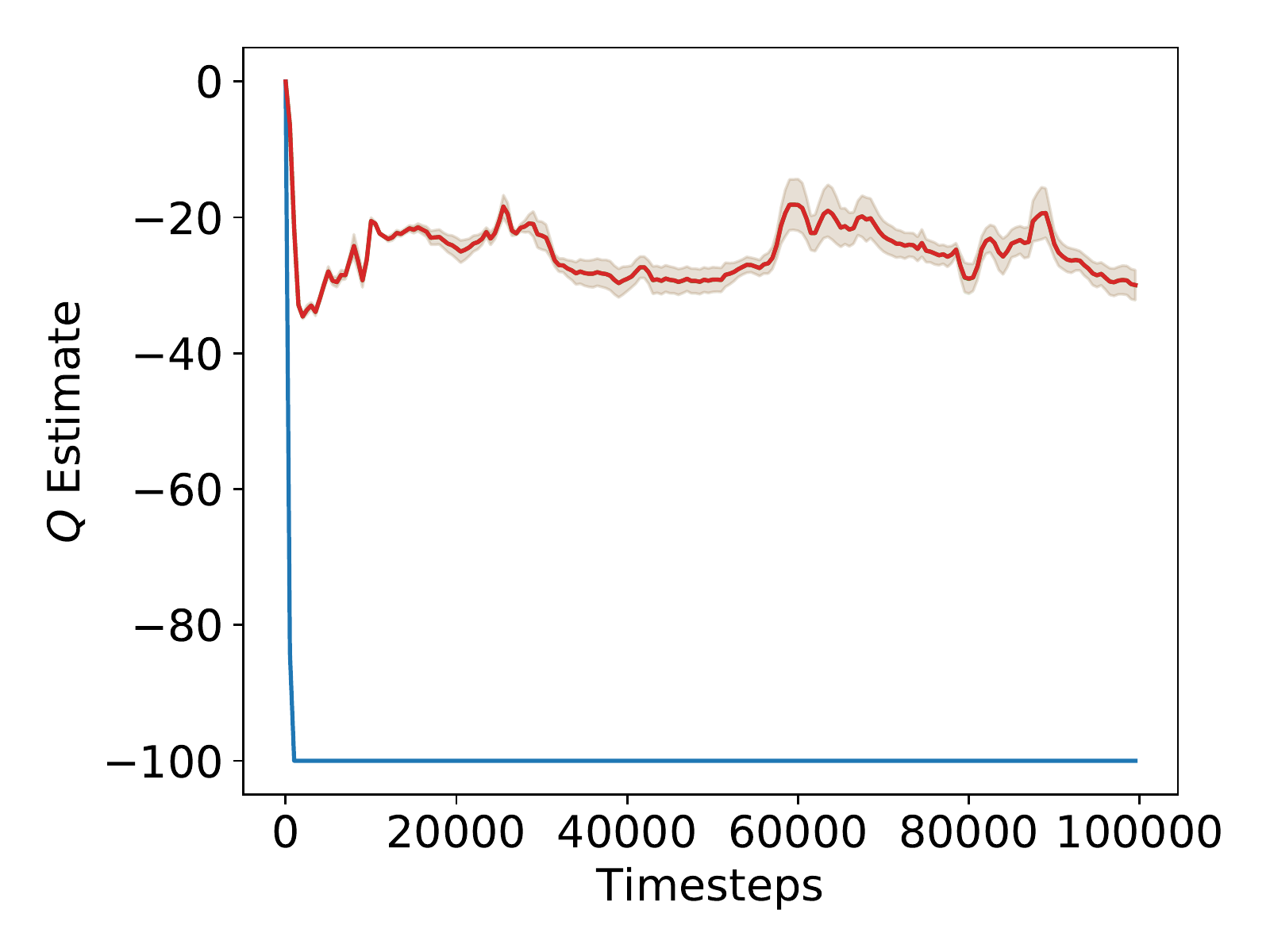}
        \label{fig:fail-scale}
    }

    \includegraphics[width=0.6\columnwidth]{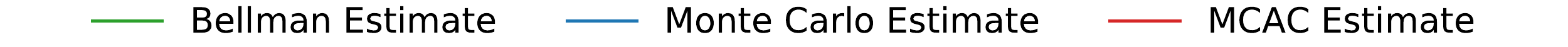}
     \caption{\new{\textbf{Estimator Scale Experiment: } Plots show mean and standard error across three random seeds of the various $Q$ estimates discussed in the paper based on uniform samples from the replay buffer for (a) successful and (b) failed trajectories. (b) appears to only have two lines because the data for Bellman estimates and \algabbr{} estimates are identical, as expected for failed trajectories.}}
    \label{fig:critic-tail}
\end{figure*}

\new{In this experiment we investigate how the scales of the various $Q$ value estimates we consider vary as the agent trains. In Figure~\ref{fig:succ-scale} we present average values for $Q$ estimates based on samples from the replay buffer for successful trajectories, while in Figure~\ref{fig:fail-scale} we present the same data for failed trajectories. The results suggest that early on, Monte Carlo estimates are much higher than Bellman estimates for successful trajectories, but in the limit these two estimates converge to similar values. This confirms our claims that \algabbr{} helps to drive up $Q$ values along demonstrator trajectories early in training. For failed trajectories, Monte Carlo estimates are uniformly as low as possible but the Bellman estimate and \algabbr{} are identical because the low Monte Carlo estimates are always cancelled out by the $\mathrm{max}$ term in the \algabbr{} estimate.}

\subsection{AWAC $Q$ Estimates}
\begin{figure*}[htb!]
\centering
    
    \includegraphics[width=0.4\columnwidth]{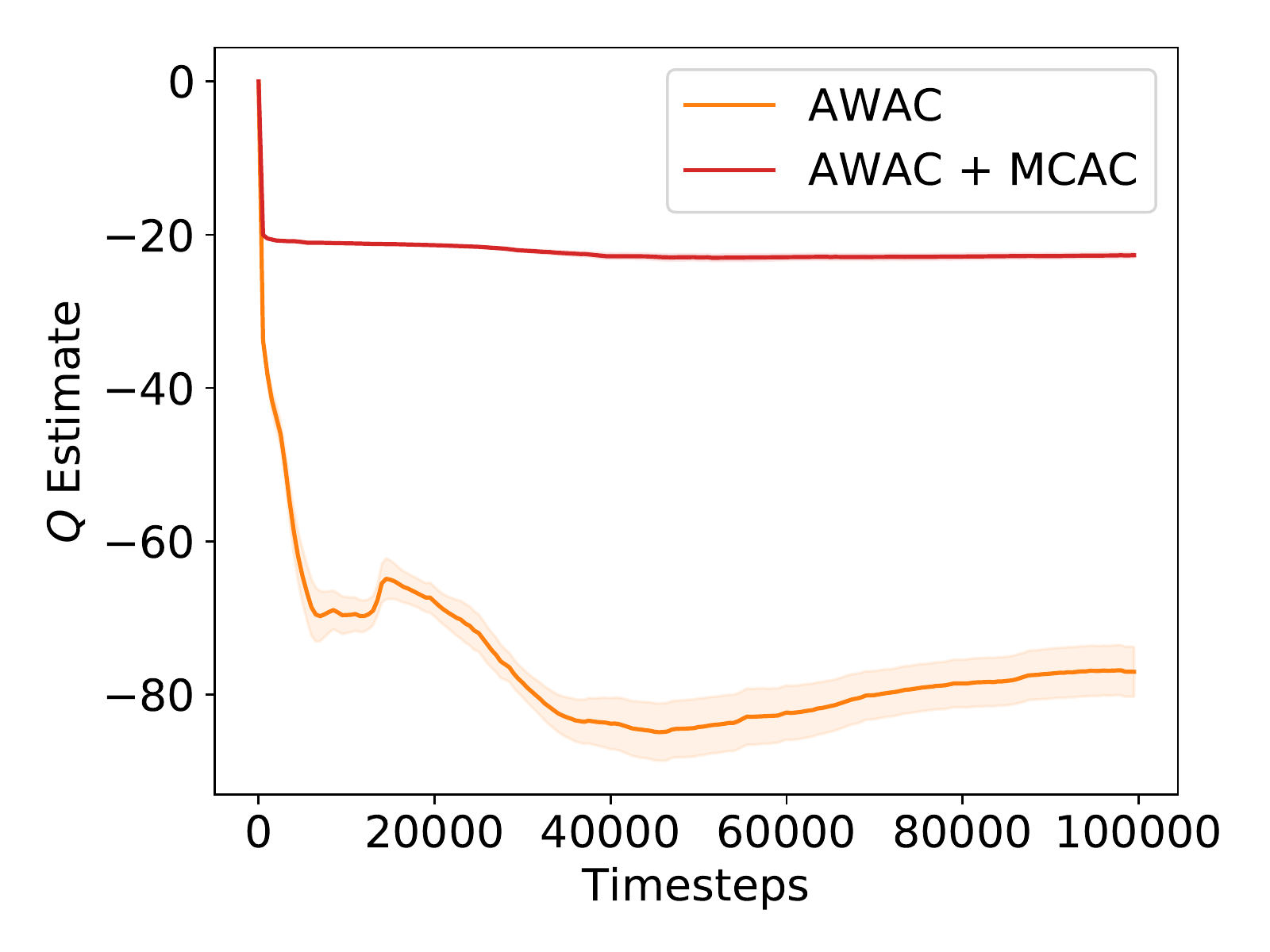}
     \caption{\new{\textbf{AWAC $Q$ Estimates: } Plots show mean and standard error across five random seeds of the $Q$ estimates from the AWAC and AWAC+\algabbr{} learners. At the $0$th timestep, each algorithm has undergone a pretraining procedure as described in Section~\ref{subsec:awac}. We see that while the original $Q$ estimate plummets after online data is added to the buffer, the version with \algabbr{} stays relatively constant.}}
    \label{fig:awac-estimates}
\end{figure*}

\new{In this section we briefly study the results of the AWAC+\algabbr{} experiment in the navigation environment, where AWAC's policy fell apart during online training while AWAC+\algabbr{} was more stable. Although there are too many factors at play to arrive at a concrete conclusion, a factor that likely differentiated the algorithms is the $Q$ value scales, presented in Figure ~\ref{fig:awac-estimates}. In these results, which show the mean $Q$ estimates computed during gradient updates for AWAC with and without \algabbr{}, we see that \algabbr{} helps to prevent the $Q$ values from crashing with the addition of online data, which they do without it. This phenomenon likely played a role in the difference between the two algorithms.}

\end{document}